\newif\ifisarxiv
\isarxivtrue

\ifisarxiv
\documentclass[11pt,a4paper]{article}
\usepackage[margin=1in]{geometry}
\else
\documentclass[11pt,draftcls,onecolumn,journal]{IEEEtran}
\fi
\usepackage{amsmath,amsfonts,amssymb,amsthm}
\usepackage{algorithmic}
\usepackage{algorithm}
\usepackage{array}
\usepackage{subcaption}
\usepackage{textcomp}
\usepackage{stfloats}
\usepackage{url}
\usepackage{verbatim}
\usepackage{graphicx}
\usepackage{cite}
\usepackage{hyperref}
\usepackage{booktabs}
\usepackage{xcolor}
\usepackage[capitalise, nameinlink]{cleveref}
\usepackage[most,skins,xparse,breakable]{tcolorbox}
\usepackage{enumitem}
\tcbset{
    boxsep=0.5mm,
    boxrule=0pt,
    colframe=white,
    arc=0mm,
    left=0.5mm,
    right=0.5mm
}

\usepackage{makecell}
\usepackage{tikz,pgfplots}
\pgfplotsset{compat=newest}
\usetikzlibrary{matrix,shapes, positioning,arrows,fit,positioning,decorations.pathreplacing}

\DeclareMathOperator{\tr}{tr}

\DeclareMathOperator{\diag}{diag}

\DeclareMathOperator{\Var}{Var}

\DeclareMathOperator{\supp}{supp}

\DeclareMathOperator{\train}{train}
\DeclareMathOperator{\test}{test}

\DeclareMathOperator{\NTK}{NTK}

\newcommand{\RR}{{\mathbb{R}}}
\newcommand{\CC}{{\mathbb{C}}}
\newcommand{\EE}{{\mathbb{E}}}
\newcommand{\NN}{{\mathcal{N}}}

\newcommand{\A}{\mathbf{A}}
\newcommand{\B}{\mathbf{B}}
\newcommand{\C}{\mathbf{C}}

\newcommand{\F}{\mathbf{F}}
\newcommand{\G}{\mathbf{G}}
\newcommand{\I}{\mathbf{I}}

\newcommand{\K}{\mathbf{K}}

\newcommand{\U}{\mathbf{U}}
\newcommand{\W}{\mathbf{W}}
\newcommand{\X}{\mathbf{X}}

\newcommand{\Z}{\mathbf{Z}}
\newcommand{\Q}{\mathbf{Q}}

\newcommand{\x}{\mathbf{x}}
\newcommand{\y}{\mathbf{y}}
\newcommand{\z}{\mathbf{z}}

\newcommand{\w}{\mathbf{w}}

\newcommand{\vv}{\mathbf{v}}
\newcommand{\uu}{\mathbf{u}}

\newcommand{\zo}{\mathbf{0}}
\newcommand{\one}{\mathbf{1}}

\newcommand{\bbeta}{\boldsymbol{\beta}}

\newcommand{\bTheta}{\boldsymbol{\Theta}}

\newcommand{\bLambda}{\boldsymbol{\Lambda}}

\newcommand{\bPhi}{\boldsymbol{\Phi}}

\definecolor{RED}{rgb}{0.7,0,0}
\definecolor{BLUE}{rgb}{0,0,0.69}
\definecolor{GREEN}{rgb}{0,0.6,0}
\definecolor{PURPLE}{rgb}{0.69,0,0.8}

\newcommand{\RED}{\color[rgb]{0.70,0,0}}
\newcommand{\BLUE}{\color[rgb]{0,0,0.69}}

\newtheorem{definition}{Definition}

\newtheorem{Theorem}{Theorem}

\newtheorem{Proposition}{Proposition}

\newtheorem{Remark}{Remark}
\newtheorem{Example}{Example}
\newtheorem{Observation}{Observation}

\begin{document}

\ifisarxiv
\title{Random Matrix Theory for Deep Learning:\\ Beyond Eigenvalues of Linear Models}
\author{
  Zhenyu Liao\\
  School of Electronic Information and Communications,\\
  Huazhong University of Science and Technology, Wuhan, China\\
  \texttt{zhenyu\_liao@hust.edu.cn}
  \and
  Michael W.\@ Mahoney\\
  ICSI, LBNL, and Department of Statistics\\
  University of California, Berkeley, USA\\
  \texttt{mmahoney@stat.berkeley.edu}
}
\date{\today}
\fi

\maketitle


\ifisarxiv
\section{Introduction}
\fi

Modern applications of Machine Learning (ML) and Artificial Intelligence (AI) aim to extract insights from high-dimensional datasets using over-parameterized ML models such as Deep Neural Networks (DNNs).
Data are typically represented as high-dimensional random vectors (of dimension $p$); numerous random vectors can be arranged into large random matrices (of size $p \times n$); and these can further be arranged into tensors.
The resulting ML models—often composed of vectors, matrices, and tensors with entrywise nonlinearities—are used for ML tasks such as classification, regression, etc., based on their low-dimensional (or even scalar) outputs.
Recently, such data matrices and ML models have begun to be studied through the lens of Random Matrix Theory (RMT).

Originating in the work of Wishart and Wigner, RMT traditionally focuses on the \emph{eigenvalue distribution} of random matrices, and it has a long history of success in fields as diverse as physics~\cite{potters2020first}, statistics~\cite{anderson2010introduction,bai2010spectral}, quantitative finance~\cite{bun2017Cleaning}, signal processing (SP), and wireless communication~\cite{tulino2004random,couillet2011random}. 
However, most existing RMT research is focused on \emph{eigenvalues} of \emph{linear} models, limiting its applicability to many modern ML models, including DNNs.
Recent advances have begun expanding RMT to address both of these constraints, substantially broadening its scope to include \emph{non-eigenvalue} analyses of \emph{nonlinear}~models.

Another axis on which to understand research in this area is that classical asymptotic statistics and concentration inequalities provide powerful tools to assess random matrices and ML models in what is often referred to as the \emph{classical regime}, i.e., when the sample size $n$ is much larger than the dimension $p$.
However, insights drawn from low-dimensional geometry—whether based on everyday (two-dimensional or three-dimensional) intuition or on theoretical results tailored to the (well-concentrating and relatively well-behaved) classical statistical regime—can become \emph{inaccurate}, and sometimes even \emph{misleading}, when applied to modern large-scale ML models.
This issue is particularly pronounced in what may be called the \emph{proportional regime}, where $n$ and $p$ are both large and comparable (i.e., when $n \sim p \gg 1$).
In this regime, a range of \emph{counterintuitive} phenomena arise, extending well beyond the well-known (and relatively well-behaved) ``curse of dimensionality.''
Perhaps the most striking example is the so-called ``double descent'' phenomenon~\cite{hastie2022Surprises,mei2021generalization}, which remains ``hidden'' under the classical assumption $n \gg p$, but which emerges prominently in the proportional regime. 

As a result, many widely-used linear and nonlinear ML models, originally designed with classical-regime intuitions in mind, exhibit markedly different behaviors in the proportional regime.
This \emph{paradigm shift} fundamentally challenges several core principles for designing ML models, especially DNNs.
Recent advances in deep learning theory—spanning deep Gaussian process~\cite{lee2018deep}, Neural Tangent Kernel~\cite{jacot2018neural}, benign overfitting~\cite{bartlett2020benign}, and double descent~\cite{hastie2022Surprises,mei2021generalization}—underscore the emerging need for novel analytic tools to study ML models in the proportional regime.

\ifisarxiv
\medskip
\fi
\emph{The goal of this paper is to provide an overview of recent advances in Random Matrix Theory (RMT) analysis for modern Deep Learning (DL), with an emphasis on going well beyond traditional eigenvalue analysis of linear models}.
We aim to highlight the \emph{novel technical challenges} that arise in the analysis of large-scale ML models such as DNNs.
In particular, ML and DL diverge from traditional RMT application domains (e.g., SP and wireless communication) in the following ways.
\begin{enumerate}
	\item \textbf{Beyond eigenvalue analysis}. 
  In SP and wireless communication, the focus is typically on eigenvalue distributions and their linear statistics (e.g., for evaluating the ergodic capacity of communication systems~\cite{tulino2004random,couillet2011random}).
	In contrast, ML and DL emphasize model performance metrics such as classification or regression errors. 
  These quantities involve \emph{both} eigenvalues and eigenvectors, alongside nonlinear transformations, derivatives, and integrations.
  As a result, a \emph{more systematic technical approach} is required to analyze these performance metrics and optimize ML model design accordingly.
  One such methodology, the \emph{Deterministic Equivalent for Resolvent}, will be discussed in detail.
	\item \textbf{Nonlinear and structured models}.
  ML and DL methods are designed to handle \emph{realistic, structured} data such as images and text.
  To extract meaningful features from these data for downstream tasks, DNNs use \emph{entrywise nonlinear} activations, multilayer architectures, and domain-specific structures like convolutional filters, weight sharing (recurrent or otherwise), and self-attention mechanisms.
  These nonlinearities and design complexities set ML and DL models apart from the simpler, typically linear, models that have long been central in traditional SP and wireless communication research.
\end{enumerate}

The reader may have noticed that \emph{asymptotic versus non-asymptotic} analysis is not listed among these differences.
Non-asymptotic approaches are indeed crucial in modern ML and AI (and they differ from the classical asymptotic focus in RMT), and recent years have seen progress in non-asymptotic RMT~\cite{vershynin2012introduction}. 
However, providing a broad, non-asymptotic overview of RMT would introduce additional technical complexity that might obscure our key message: \emph{RMT can extend well beyond the eigenvalues of linear models, making it far more relevant to a wide variety of realistic ML settings}.

\ifisarxiv
\medskip
\fi

In this overview, we focus on analyzing a large (D)NN model $\mathcal{M}_{\phi}(\X;\bTheta)$, where $\X$ represents random input data, $\bTheta$ represents the model's weight parameters, and $\phi$ is an entrywise nonlinear (or, as a special case, linear) function. 
Formal definitions of a linear model, a nonlinear single-hidden-layer NN model, and a nonlinear DNN model can be found in Definitions~\ref{def:noisy_linear_model}, \ref{def:single-layer-NN}, and \ref{def:DNN}, respectively. 
As is typical in ML, our goal is to evaluate $\mathcal{M}_{\phi}(\X;\bTheta)$ through some \emph{scalar performance metric} $f(\cdot)$, such as the mean squared error (MSE), classification error, and other commonly used metrics such as cross-entropy and perplexity, in the \emph{proportional regime}, when the sample size $n$, the input dimension $p$, and the number of model parameters $d$ are all large and comparable.

Analyzing $f\left(\mathcal{M}_{\phi}(\X;\bTheta) \right)$ poses significant technical challenges for the following reasons.
\begin{enumerate}
  \item \textbf{High-dimensionality of $\X $ and $ \bTheta$.} 
  The input data $\X$ and (randomly initialized) weight parameters $\bTheta$ are often modeled as large random matrices. 
  Consequently, the statistical behavior of $f\left(\mathcal{M}_{\phi}(\X;\bTheta) \right)$ may depend on the distribution of $\X$ and $\bTheta$ in a non-trivial fashion.
  \item \textbf{Analysis of eigenspectral functional.} 
  When a training procedure is introduced, the model parameters $\bTheta$ become functions of $\X$, depending on its eigenvalues and eigenvectors, even for linear models (see \Cref{def:noisy_linear_model} and the subsequent discussion for a concrete example).
  \item \textbf{Nonlinearity in the model.} 
  DNN models often incorporate \emph{entrywise nonlinear} activations $\phi$ and multilayer architectures, which adds another layer of complexity to their analysis.
\end{enumerate}
As the main technical contribution and key takeaway of this paper, we can address all three challenges in the analysis of large-scale \emph{random nonlinear} NN models by introducing their corresponding \emph{High-dimensional Equivalent}.
This important notion is formally stated below.

\begin{tcolorbox}[breakable]
\begin{definition}[\textbf{High-dimensional Equivalent}]
\label{def:HiE}
Let $\mathcal{M}_{\phi}(\X) \in \RR^{p \times n}$ be a (nonlinear) random matrix model that depends on a random matrix $\X \in \RR^{p \times n}$ and function $\phi \colon \RR \to \RR$ that applies entrywise. 
Let $f \left( \mathcal{M}_{\phi}(\X) \right)$ be a scalar observation of $\mathcal{M}_{\phi}(\X)$ for some $f \colon \RR^{p \times n} \to \RR$.
We say that $\tilde{\mathcal{M}}_\phi(\X)$ (random or deterministic) is a \emph{High-dimensional Equivalent} of $\mathcal{M}_{\phi}(\X)$ with respect to $f(\cdot)$ if
\begin{equation}\label{eq:HiE}
  f(\mathcal{M}_{\phi}(\X)) - f(\tilde{\mathcal{M}}_{\phi}(\X)) \to 0,
\end{equation}
in probability or almost surely as $n,p \to \infty$ with $p/n \to c \in (0,\infty)$.
We denote this relation as $\mathcal{M}_{\phi}(\X) \overset{f}{\leftrightarrow} \tilde{\mathcal{M}}_{\phi}(\X)$, or simply $\mathcal{M}_{\phi}(\X) \leftrightarrow \tilde{\mathcal{M}}_{\phi}(\X)$ when $f$ is clear from context.
\end{definition}
\end{tcolorbox}

More precisely, we show the following.
\begin{enumerate}
    \item 
    In the absence of nonlinearities, for example when $\mathcal{M}_{\phi}(\X) = \X$, the scalar functional $f(\X)$ of a linear random matrix model $\X$ concentrates around its expectation $f(\X) \simeq \EE[f(\X)]$ for some specific $f(\cdot)$ of interest for ML purposes, see for example \Cref{rem:scalar_func}, and it can be assessed through its \emph{Deterministic Equivalent} $f(\tilde\X)$. 
    See \Cref{def:DE}, which is a special case of \Cref{def:HiE}.
    \item 
    For scalar eigenspectral functionals as defined in \Cref{def:spectral_functional}, we introduce in \Cref{theo:spectral_func_contour} a \emph{Deterministic Equivalent for Resolvent} framework, and we show that this provides a unified approach to eigenspectral functionals of large random matrices.
    \item 
    For nonlinear models, we consider two different scaling regimes (see \Cref{def:two_scaling_regimes}); and we show that in these two regimes the nonlinear models $\mathcal{M}_{\phi}(\X)$ can be linearized to yield a \emph{Linear Equivalent} (\Cref{def:Hi-LE}, another special case of \Cref{def:HiE}). 
    The resulting linearized model can then be tackled much like a linear model.
\end{enumerate}
An overview of these concepts and their applications to linear, shallow nonlinear, and deep nonlinear models is provided in \Cref{fig:appetizer}.
\begin{figure}[!tb]
\centering
\usetikzlibrary{decorations.pathreplacing,positioning,shadows,calc}
\tikzset{
  Def1/.style={
    rectangle,
    draw=#1!70!black,
    top color=#1!20!white,
    bottom color=#1!50!white,
    drop shadow,
    rounded corners=3pt
  },
  Def2/.style={
    rectangle,
    draw=#1!70!black,
    top color=#1!20!white,
    bottom color=#1!50!white,
    drop shadow
  },
  Remark/.style={rectangle,minimum width=#1,minimum height=6mm},
  Frame/.style={rectangle,draw,minimum width=#1,minimum height=10mm}
}

{
\scriptsize
\begin{tikzpicture}[>=stealth]
  \node (Def19) [Def1=blue]{\parbox{2.5cm}{\centering \textbf{Linear model}\\ \textcolor{darkgray}{(\Cref{def:noisy_linear_model})}}};
  \node (Def23) [Def1=blue,below=4mm of Def19]{\parbox{2.5cm}{\centering \textbf{Nonlinear shallow NN} \textcolor{darkgray}{(\Cref{def:single-layer-NN})}}};
  \node (Def31) [Def1=blue,below=4mm of Def23]{\parbox{2.5cm}{\centering \textbf{Nonlinear DNN} \\ \textcolor{darkgray}{(\Cref{def:DNN})}}};
  \node (Remark21) [Remark=2cm,right=5mm of Def19] {\Cref{rem:scaling_in_sample}};
  \node (Remark27) [Remark=2cm,right=5mm of Def23] {\Cref{rem:scaling_train_NN}};
  \node (Remark33) [Remark=4.5cm,right=5mm of Def31] {\Cref{rem:deep-versus-shallow}};
  \node (Remark22) [Remark=2cm,right=5mm of Remark21] {\Cref{rem:double_descent_out-of-sample}};
  \node (Remark28) [Remark=2cm,right=5mm of Remark27] {\Cref{rem:double_descent_test_NN}};
  \node (Remark34) [Remark=2cm,right=5mm of Remark33] {\Cref{rem:DNN_dynamics}};
  \node (F1)[Frame=2cm,above=5mm of Remark21]{\parbox{2cm}{\centering \textbf{Scaling law of training error}}};
  \node (F2)[Frame=2cm,above=5mm of Remark22]{\parbox{2cm}{\centering \textbf{Double descent test error}}};
  \coordinate (A) at (Remark34.west |-Remark28.south);
  \coordinate (B) at (Remark34.west |-Remark22.north);
  \coordinate (C) at (Remark34.east |-Remark22.north);
  \coordinate (D) at (Remark34.east |-Remark28.south);
  \node (F3) at (Remark34|-F2)[Frame=2cm]{\parbox{2cm}{\centering \textbf{Learning dynamics}}};
  \foreach \x in {21,22,27,28,33,34}
  {
    \draw[densely dashed,semithick](Remark\x.south west)[rounded corners=7pt]--(Remark\x.north west)[sharp corners]--(Remark\x.north east)[rounded corners=7pt]--(Remark\x.south east)[sharp corners]--cycle;
  }
  \draw[densely dashed,semithick](A)[rounded corners=7pt]--(B)[sharp corners]--(C)[rounded corners=7pt]--(D)[sharp corners]--cycle;
  \draw[white,text=black](A)--(C)node[midway]{\Cref{rem:dynamics_NN}};
  \coordinate (F) at ($(Def19.north west)!0.5!(Def23.south west)$);
  \coordinate (G) at ($(Def23.north west)!0.5!(Def31.south west)$);
  \node (Def3) at ([xshift=-2cm]F)[Def2=red]{\parbox{2.5cm}{\centering \textbf{Deterministic Equiv.\@} \\ \textcolor{darkgray}{(\Cref{def:DE})}}};
  \node (Def14) at ([xshift=-2cm]G)[Def2=red]{\parbox{2.5cm}{\centering \textbf{Linear Equivalent} \\ \textcolor{darkgray}{(\Cref{def:Hi-LE})}}};
  \draw[ultra thick,->,darkgray](Def3)--(Def19);
  \draw[ultra thick,->,darkgray](Def3)--(Def23);
  \draw[ultra thick,->,darkgray](Def14)--(Def31);
  \draw[ultra thick,->,darkgray](Def14)--(Def23);
  \coordinate (H) at ($(Def3.north west)!0.5!(Def14.south west)$);
  \node (Def1) at ([xshift=-2cm]H)[Def2=green]{\parbox{2.5cm}{\centering \textbf{High-dimensional} \\ \textbf{Equivalent} \\ \textcolor{darkgray}{(\Cref{def:HiE})}}};
  \draw[ultra thick,->,darkgray](Def1)--(Def3);
  \draw[ultra thick,->,darkgray](Def1)--(Def14);
  \coordinate (E) at (Def31.south-|Def14.east);
  \draw[thick,RED,densely dashed](Def1.west|-F1.north)rectangle([xshift=3mm]E)node[at start,below right]{\large\bfseries Proposed RMT framework};
\end{tikzpicture}
}
\caption{{ Overview of this paper, summarizing major concepts and results and where to find them. }}
\label{fig:appetizer}
\end{figure}

\ifisarxiv
\paragraph{Organization of this paper.}
The remainder of this paper is organized as follows. 
In \Cref{sec:DE_for_resolvent}, we start with a discussion on the concentration behavior of scalar observations of high-dimensional random vectors and matrices, and introduce the \emph{Deterministic Equivalent for Resolvent} framework, which provides unified access to eigenspectral functionals of large random matrices.
To address the challenges posed by nonlinearity, we present linearization techniques in \Cref{sec:high_dimensional_linearization} that allow us to analyze high-dimensional nonlinear models.
\Cref{sec:linear_model} and \Cref{sec:shallow_NN} introduce the linear model and the single-hidden-layer nonlinear NN model, respectively, and demonstrate how our framework can be applied to assess their performance in the proportional regime.
We then extend the analysis to \emph{deep} networks in \Cref{sec:DNN}, moving beyond the single-hidden-layer setting. 
Finally, \Cref{sec:conclusion} concludes the paper with a summary of our contributions and a discussion of future perspectives. 
Additional technical details and proofs are provided in the appendices.
\fi

\ifisarxiv
\section{Deterministic Equivalent for Resolvent: A Unified Framework for Modern RMT Beyond Eigenvalues}\label{sec:DE_for_resolvent}
\fi

In this section, we focus on eigenspectral functionals of linear models (with $\phi(t) = t $ in \Cref{def:HiE}), and we introduce the \emph{Deterministic Equivalent for Resolvent} framework as a unified analysis approach.

\ifisarxiv
\subsection{Concentration of scalar functionals of large random vectors and matrices}
\fi

We begin with a simple but illustrative observation about the behavior of random vectors in one and multiple dimensions.
\begin{Observation}[\textbf{``Concentration'' versus ``non-concentration'' around the mean}]
\label{rem:non_determ}\normalfont
Consider two independent random vectors $\x = [x_1, \ldots, x_n]^\top$ and $ \y = [y_1, \ldots, y_n]^\top \in \RR^n$, with i.i.d.\@ entries of zero mean and unit variance. 
We have the following observations.
\begin{enumerate}
    \item In the one-dimensional case with $n = 1$, we have $\Pr ( |x - 0| > t ) \leq t^{-2}$ and $\Pr ( |y - 0| > t ) \leq t^{-2}$ by Markov's inequality, so that one-dimensional random variables ``concentrate'' around their means.
    \item In the multi-dimensional case with $n \geq 1$, we have $\EE[\| \x - \zo \|_2^2] = \EE[ \x^\top \x ] = \tr (\EE[\x \x^\top])  = n$ and $\EE[ \| \x - \y \|_2^2 ] = \EE[ \x^\top \x + \y^\top \y] = 2 n$.
    Thus, for $n \gg 1$, the expected Euclidean distance between $\x$ and its mean $\zo$ is large:
    high-dimensional random vectors do \emph{not} ``concentrate'' around their means.
\end{enumerate}
\end{Observation}
From \Cref{rem:non_determ}, we see that while one-dimensional random variables typically remain close to their means, as shown in \Cref{fig:low-dimen_vec}, this is \emph{no longer true} for high-dimensional random vectors with $n \gg 1$. 
In that setting, any two independent draws $\x, \y$ of random vectors in $\RR^n$ are approximately orthogonal to each other, and form, together with the origin (which is also the mean) an isosceles right triangle, implying that any high-dimensional vector is \emph{not} close to its mean.
This counterintuitive ``concentration versus non-concentration around the mean'' phenomenon is depicted (informally, since the depiction of this fundamentally high-dimensional phenomenon is being shown on a two-dimensional piece of paper) in \Cref{fig:low-high-dimen_vec}.

\begin{figure}[!tb]
\centering
\begin{subfigure}[t]{0.48\linewidth}
\centering
\begin{tikzpicture}[scale=.8]
\centering
\renewcommand{\axisdefaulttryminticks}{4} 
\pgfplotsset{every axis legend/.append style={cells={anchor=west},fill=white, at={(0.98,0.98)}, anchor=north east, font=\scriptsize }}
    \begin{axis}[
      axis y line*=left,
      height=.7\linewidth,
      xmin=-3,xmax=3,
      ymax=.5,
      xtick={-3,0,3},
      bar width=4pt,
      ymajorgrids=false,
      grid=major,
      ylabel= { {Histogram} }
      ]
      \addplot+[ybar,mark=none,draw=white,fill=BLUE!50!white,area legend] coordinates{
      (-2.896552, 0.002360)(-2.689655, 0.016520)(-2.482759, 0.023600)(-2.275862, 0.037760)(-2.068966, 0.054281)(-1.862069, 0.068441)(-1.655172, 0.101481)(-1.448276, 0.132161)(-1.241379, 0.169922)(-1.034483, 0.228923)(-0.827586, 0.287923)(-0.620690, 0.370524)(-0.413793, 0.346924)(-0.206897, 0.351644)(-0.000000, 0.382324)(0.206897, 0.398844)(0.413793, 0.363444)(0.620690, 0.361084)(0.827586, 0.245443)(1.034483, 0.261963)(1.241379, 0.195882)(1.448276, 0.160482)(1.655172, 0.082601)(1.862069, 0.054281)(2.068966, 0.044840)(2.275862, 0.044840)(2.482759, 0.011800)(2.689655, 0.011800)(2.896552, 0.009440)(3.103448, 0.000000)};
      \addplot[smooth,samples=300,domain=-3:3,line width=1.5pt,RED] {  exp(-x^2/2)/sqrt(2*pi) };
    \end{axis}
  \end{tikzpicture}
\caption{{ ``Concentration'' around the mean for one-dimensional random vectors }}
\label{fig:low-dimen_vec}
\end{subfigure}
\hfill
\begin{subfigure}[t]{0.48\linewidth}
\centering
\begin{tikzpicture}[scale=.8]
\centering
  \shade[ball color=white,opacity=0.25](0,0)circle (2);
  \draw(0,0) circle(2);
  \draw[thick,BLUE](-2,0) arc (180:360:2 and 0.6);
  \draw[dashed,BLUE] (2,0) arc (0:180:2 and 0.6);
  \fill[fill=black] (0,0) circle (1pt);
  \draw[RED,thick,->] (0,0) -- node[above]{$\x$} (2,0);
  \draw[RED,thick,->] (0,0) -- node[right]{$\y$} (0,2);
\draw node[below,yshift=0.1cm] at (0,0) {\footnotesize{$\EE[\x] = \EE[\y] = \zo_n$}};
  \draw (0,0.2) -- (0.2,0.2) -- (0.2,0);
  \draw node at (0.3,0.3) {\footnotesize{$\approx$}};
  \draw node[right] at (2,0) {$\| \x \|_2 \approx \sqrt n$};
  \draw node[above] at (0,2) {$\| \y \|_2 \approx \sqrt n$};
\end{tikzpicture}
\caption{{ ``Non-concentration'' around the mean for multi-dimensional random vectors }}
\label{fig:high-dimen_vec}
\end{subfigure}
\caption{ { Visualization of ``concentration'' versus ``non-concentration'' around the mean for one-dimensional~versus~multi-dimensional random vectors in \Cref{rem:non_determ}. } }  
\label{fig:low-high-dimen_vec}
\end{figure} 

While multi-dimensional random vectors do not ``concentrate'' around their means (see \Cref{rem:non_determ} and \Cref{fig:high-dimen_vec}), their \emph{scalar} functionals often do.
Concretely, for a \emph{scalar} observation function $f \colon \RR^n \to \RR$ and random vector $\x \in \RR^n$, we typically have
\begin{equation}\label{eq:f(x)_summary_Chap1}
  f(\x) - \EE[f(\x)] \to 0,
\end{equation}
in probability or almost surely as $n \to \infty$. 
A basic example is the linear function $f(\x) = \one_n^\top \x/n = \frac1n \sum_{i=1}^n x_i$.
By the Law of Large Numbers (LLN) and the Central Limit Theorem (CLT), we have that $f(\x)$ stays close to its expectation $ \EE[f(\x)]$, up to an $O(n^{-1/2})$ deviation, with high probability.

Matrices, as a natural extension of vectors, exhibit similar behaviors.
For a random matrix $\X \in \RR^{p \times n}$ in the proportional regime with $n,p$ both large,
we have the following.
\begin{enumerate}
  \item 
  Just as in \Cref{rem:non_determ} for vectors, the random matrix $\X$ does \emph{not} concentrate, e.g., in a spectral norm sense; for instance, $\| \X - \EE[\X] \|_2 \not \to 0$ as $n,p \to \infty$ together. 
  \item 
  At the same time, similar to \eqref{eq:f(x)_summary_Chap1}, scalar (e.g., eigenspectral) functionals $f \colon \RR^{p\times n} \to \RR$ of the random matrix $\X$ \emph{do} concentrate; namely, $f(\X) - \EE[f(\X)] \to 0$ as $n,p \to \infty$. 
\end{enumerate}

\noindent
Hence, when the focus is on scalar functionals $f(\cdot)$ of a random matrix $\X$—common in ML and related fields—it becomes possible to find a deterministic matrix $\tilde\X$ that ``mimics'' $\X$, but \emph{only} through the observation function $f(\cdot)$.
We refer to this deterministic matrix $\tilde\X$ as a \emph{Deterministic Equivalent} of $\X$. 

\begin{tcolorbox}[breakable]
\begin{definition}[\textbf{Deterministic Equivalent},~{\cite{hachem2007deterministic},~\cite[Section~2]{couillet2022RMT4ML}}]\label{def:DE}
A \emph{Deterministic Equivalent} is a special case of the \emph{High-Dimensional Equivalent} in \Cref{def:HiE} for \emph{deterministic} $\tilde \X$, applied to a linear random matrix model $\mathcal{M}_{\phi}(\X) = \X$.
We denote 
\begin{equation}
f(\X) - f(\tilde\X) \to 0~\text{as}~n,p\to\infty \quad \Leftrightarrow \quad \X \overset{f}{\leftrightarrow} \tilde\X ~\text{or simply}~\X \leftrightarrow \tilde\X.
\end{equation}
\end{definition}
\end{tcolorbox}
\begin{Remark}[\textbf{Commonly-used scalar functionals}]\normalfont\label{rem:scalar_func}
When dealing with scalar eigenspectral functionals of large random matrices (see \Cref{def:spectral_functional}),  a common approach is to evaluate trace forms $\tr(\Q \A)/n$ (that is connected to the Stieltjes transform in \Cref{def:resolvent}) and bilinear forms $\mathbf{a}^\top \Q \mathbf{b}$ (that is connected to eigenspectral functionals via contour integration per \Cref{theo:spectral_func_contour}) of the resolvent matrix $\Q \in \RR^{n \times n}$ (see \Cref{def:resolvent}), where $\A \in \RR^{n \times n}, \mathbf{a}, \mathbf{b} \in \RR^n$ have bounded spectral and Euclidean norms as $n \to \infty$.
These objects, and their role in characterizing eigenspectral functionals of large random matrices, will be discussed in more detail in the following section.
\end{Remark}

\ifisarxiv
\subsection{A unified framework for scalar eigenspectral functionals via the resolvent}
\fi

Classical RMT primarily focuses on eigenvalue distributions of large random matrices.
\begin{tcolorbox}[breakable]
\begin{definition}[\textbf{Empirical Spectral Distribution/Measure, ESD or ESM}~\cite{bai2010spectral}]\label{def:ESD}
Let $\X \in \RR^{n \times n}$ be a symmetric matrix with eigenvalues $\lambda_1(\X),\ldots,\lambda_n(\X)$.
The Empirical Spectral Distribution/Measure $\mu_\X$ of $\X$ is defined as the normalized counting measure of its eigenvalues.
That is, $\mu_\X \equiv \frac1n \sum_{i=1}^n \delta_{\lambda_i(\X)}$, where $\delta_x$ is the Dirac measure at $x$. 
\end{definition}
\end{tcolorbox}
\noindent
Note from \Cref{def:ESD} that the ESD $\mu_\X$ of a symmetric matrix $\X$ is a probability measure.

For ML purposes, one often evaluates a random matrix $\X \in \RR^{n \times n}$ through a scalar (performance) metric that depends on both its eigenvalues and eigenvectors.
Given a function $f \colon \RR \to \RR$, we define the associated \emph{matrix function} as $\U_\X f(\bLambda_\X)\U_\X^\top$, where $\X=\U_\X \bLambda_\X\U_\X^\top$ is the eigen-decomposition of $\X$ and $f$ is applied entrywise to the diagonal entries of $\bLambda_\X$.
The bilinear forms involving such matrix functions are referred to as the \emph{scalar eigenspectral functionals} of $\X$ throughout this paper.

\begin{tcolorbox}[breakable]
\begin{definition}[\textbf{Scalar eigenspectral functional}]
\label{def:spectral_functional}
Let $\X \in \RR^{n \times n}$ be a symmetric matrix with eigen-decomposition $\X=\U_\X \bLambda_\X\U_\X^\top = \sum_{i=1}^n \lambda_i(\X) \uu_i \uu_i^\top$, for $\U_\X=[\uu_1,\ldots,\uu_n]\in\RR^{n\times n}$ and $\bLambda_\X=\diag\{\lambda_1(\X),\ldots,\lambda_n(\X)\}$. 
We say
\begin{equation}
  f(\X) = \frac1{|\mathcal{I}|} \sum_{i \in \mathcal{I} \subseteq \{1, \ldots, n \} } f(\lambda_i(\X)) \mathbf{a}^\top \uu_i \uu_i^\top \mathbf{b},
\end{equation}
is a \emph{scalar eigenspectral functional} of $\X$, for $\mathbf{a}, \mathbf{b} \in \RR^n$ and the index set $\mathcal{I}$ (of cardinality $|\mathcal{I}|$) that contains some or all of the eigenvalue-eigenvector pairs $(\lambda_i(\X),\uu_i)$ of $\X$.
\end{definition}
\end{tcolorbox}

\noindent
Note that the vectors $\mathbf{a}, \mathbf{b} \in \RR^n$ may or may not depend on $\X$.
Below are two special cases of scalar eigenspectral functionals defined in \Cref{def:spectral_functional}.
\begin{enumerate}
    \item \textbf{Linear Spectral/Eigenvalue Statistics} (LSS or LES) $f_\X$ of $\X$, see~\cite{lytova2009central,bai2004clt}. 
    They are the averaged statistics of the eigenvalues $\lambda_1(\X), \ldots, \lambda_n(\X)$ of $\X$ through $f \colon \RR \to \RR$, namely
    \begin{equation}
      f_\X = \frac1n \sum_{i=1}^n f( \lambda_i(\X) ) = \int f(t) \mu_\X(dt),
    \end{equation}
    which is a Lebesgue integral with respect to $\mu_\X$ the ESD of $\X$ as in \Cref{def:ESD}.
    One obtains $f_\X$ from \Cref{def:spectral_functional} by setting $\mathbf{a} = \mathbf{b} = \uu_i$.
    \item \textbf{Projection} $\vv^\top\uu_i$ \textbf{of eigenvectors} $\uu_i$ onto a deterministic vector $\vv \in \RR^n$. 
    Here, $\uu_i$ is the $i^{th}$ eigenvector of $\X$. 
    One obtains this from \Cref{def:spectral_functional} by setting $\mathbf{a} = \mathbf{b} = \vv$ and $f(t) = 1$.
\end{enumerate}
Note that both the ESD and LSS do \emph{not} contain eigenvector information, but clearly the projection of eigenvectors does.
Consequently, the scalar eigenspectral functionals in \Cref{def:spectral_functional}, along with the proposed analysis framework in \Cref{theo:spectral_func_contour} below, extend beyond classical RMT objects of interest, incorporating both eigenvalue and eigenvector information.
As we shall see, these quantities naturally arise in analyzing the convergence and generalization of linear and nonlinear, shallow and deep NNs.

These and other scalar eigenspectral functionals in \Cref{def:spectral_functional} can be evaluated using the resolvent matrix and the Stieltjes transform, defined as follows.

\begin{tcolorbox}[title = \textbf{\textsf{Resolvent and Stieltjes transform}}]
\begin{definition}[\textbf{Resolvent and Stieltjes transform}]
\label{def:resolvent}
Let $\X \in \RR^{n \times n}$ be a symmetric matrix. 
Its resolvent $\Q_\X(z)$ is defined for any $z \in \CC$ that is not an eigenvalue of $\X$, as
\begin{equation}\label{eq:def-resolvent}
    \Q_\X(z) \equiv \left( \X - z \I_n \right)^{-1} \in \CC^{n \times n}.
\end{equation}
Its normalized trace satisfies, for $\lambda_1(\X),\ldots,\lambda_n(\X)$ the eigenvalues and $\mu_\X$ the ESD of $\X$ in \Cref{def:ESD},
\begin{equation}\label{eq:def_ST}
  \frac1n \tr \Q_\X(z) = \frac1n \sum_{i=1}^n \frac1{ \lambda_i(\X) - z } = \int \frac{\mu_\X(dt)}{t - z} \equiv m_{\mu_\X}(z),
\end{equation}
where $m_{\mu_\X}(z), z \in \CC \setminus \supp(\mu_\X)$ is the \emph{Stieltjes transform} of $\mu_\X$.
\end{definition}
\end{tcolorbox}

\begin{Remark}[\textbf{On the Stieltjes transform},~{\cite[Proposition~6]{pastur2003Matrices}~and~\cite[Proposition 2.1.2]{pastur2011eigenvalue}}]
\label{rem:ST}\normalfont
Let $m_\mu(z)$ be the Stieltjes transform of a probability measure $\mu$.
Then the following holds.
\begin{enumerate}
  \item The function $m_\mu$ is analytic on $\CC \setminus \RR$ and $m_\mu(\bar z) = \overline{m_\mu(z)}$, for $\bar z$ the complex conjugate of $z \in \CC$.\label{item:ST_1}
  \item $\Im[ z ] \cdot \Im[ m_\mu(z) ] > 0$ for $\Im[z] \neq 0$ and $\lim_{y\to \infty} - \jmath y m_\mu(\jmath y) = 1$, where $\jmath$ is the imaginary unit.
  \label{item:ST_2}
  \item If $m$ is a complex function satisfying items~\ref{item:ST_1}~and~\ref{item:ST_2}, then there exists a probability measure $\mu$ such that $m$ is the Stieltjes transform of $\mu$.
  \item \textbf{Inverse Stieltjes transform}: For $a,b$ continuity points of the probability measure $\mu$, then $\mu([a,b]) = \frac1{\pi} \lim_{y \downarrow 0} \int_{a}^b \Im [ m_\mu(x + \jmath y) ] dx$.
\end{enumerate}
\end{Remark}

Note that the resolvent is the more basic object, as it contains both eigenvalue and eigenvector information, while the Stieltjes transform uses the trace to remove eigenvector information.

\begin{tcolorbox}[breakable]
\begin{Theorem}[\textbf{Scalar eigenspectral functional via contour integration}]
\label{theo:spectral_func_contour}
Let $\X \in \RR^{n \times n}$ be a symmetric matrix with resolvent $\Q_\X(z) = (\X - z \I_n)^{-1}$ as in \Cref{def:resolvent}, and let $f\colon \CC \to \CC$ be analytic in a neighborhood of a positively (i.e., counterclockwise) oriented simple closed contour $\Gamma_{\mathcal I}$.
Suppose $\Gamma_{\mathcal I}$ encloses the eigenvalues $\lambda_i(\X)$ of $\X$ whose indices are in the set $\mathcal I$ and only these ones (i.e., it leaves the other eigenvalues \emph{outside}).
Then, for the scalar eigenspectral functional $f(\X)$ of $\X$ in \Cref{def:spectral_functional}, we have
\begin{equation}\label{eq:spectral_func_contour}
     f(\X) \equiv \frac1{|\mathcal{I}|} \sum_{i \in \mathcal{I} \subseteq \{1, \ldots, n \} } f(\lambda_i(\X)) \mathbf{a}^\top \uu_i \uu_i^\top \mathbf{b} = 
     -\frac1{2\pi{\jmath} |\mathcal{I}|} \oint_{\Gamma_{\mathcal I}} f(z) \mathbf{a}^\top \Q_\X(z) \mathbf{b}\,dz.
\end{equation}
\end{Theorem}
\end{tcolorbox}
\begin{proof}[Proof of \Cref{theo:spectral_func_contour}]
Given the eigen-decomposition $\X= \sum_{i=1}^n \lambda_i(\X) \uu_i \uu_i^\top$, it follows from \Cref{def:resolvent} that the resolvent admits the decomposition $\Q_\X(z) = \sum_{i=1}^n \frac{\uu_i\uu_i^\top}{\lambda_i(\X)-z}$.
Applying Cauchy's integral formula (\Cref{theo:cauchy-integral}) to this expression yields the desired result.
\end{proof}
For completeness, we state the Cauchy integral formula used in the proof of \Cref{theo:spectral_func_contour}.
\begin{tcolorbox}[breakable]
\begin{Theorem}[\textbf{Cauchy's integral formula}]
\label{theo:cauchy-integral}
Let $\Gamma \subset \CC$ be a positively (i.e., counterclockwise) oriented simple closed curve, and let $f(z)$ be a complex function analytic in a region containing $\Gamma$ and its interior.
Then, if $z_0 \in \CC$ is enclosed by $\Gamma$, $f(z_0) = -\frac1{2\pi {\jmath}} \oint_{\Gamma} \frac{f(z)}{z_0-z}\,dz$ and $ 0$ otherwise.
\end{Theorem}
\end{tcolorbox}

\noindent
\Cref{theo:spectral_func_contour} thus provides a means to evaluate scalar eigenspectral functionals (including ESD, LSS, eigenvector projections and beyond as in \Cref{def:spectral_functional}) via contour integration. 
When $\X$ is a large random matrix, its resolvent $\Q_\X(z)$ is also random, and we expect that $\mathbf{a}^\top \Q_\X(z) \mathbf{b}$ and the scalar eigenspectral functionals $f(\X)$ (per \Cref{theo:spectral_func_contour}) concentrate around their means, allowing us to apply the Deterministic Equivalent for Resolvent approach in \Cref{def:DE}.

\ifisarxiv
\section{High-dimensional Linearization: RMT for Nonlinear Models}\label{sec:high_dimensional_linearization}
\fi

In this section, we address the technical challenge of nonlinearity in ML models by introducing the concept of \emph{High-dimensional Linearization} (which will permit us to define the notion of a \emph{Linear Equivalent}, a specific instance of a High-Dimensional Equivalent in \Cref{def:HiE}). 

To start, we examine two fundamental scaling regimes—the Law of Large Numbers (LLN) and the Central Limit Theorem (CLT) regime—and discuss how different linearization approaches apply in each.

\begin{tcolorbox}[breakable]
\begin{definition}[\textbf{Two scaling regimes}]
\label{def:two_scaling_regimes}
Consider a scalar functional $f (\x)$ of a high-dimensional random vector $\x \in \RR^n$, via an observation function $f \colon \RR^n \to \RR$. 
We distinguish two \emph{scaling regimes}.
\begin{enumerate}
  \item \textbf{LLN regime}: this holds when $f(\x)$ exhibits a \emph{LLN-type concentration}, strongly concentrating around its mean $\EE[f(\x)]$, and its distribution function becomes \emph{degenerate}; that is, it holds when $f(\x) - \EE[f(\x)] \to 0$ in probability or almost surely, as $n \to \infty$.
  \item \textbf{CLT regime}: this holds when $f(\x)$ exhibits a \emph{CLT-type concentration}, remaining random and maintaining a \emph{non-degenerate} distribution function; that is, it holds when $f(\x) - \EE[f(\x)] \to \NN(0,1) $ in distribution, as $n \to \infty$. 
\end{enumerate}
\end{definition}
\end{tcolorbox}
\noindent
Below is an example of two \emph{nonlinear} functionals $\phi(f(\x))$ of $\x \in \RR^n$ in the LLN and CLT regime. 
\begin{Example}[\textbf{Nonlinear objects in two scaling regimes}]
\label{example:LLN_CLT}
Let $\x \in \RR^n$ be a random vector such that $\sqrt n\x$ has i.i.d.\@ standard Gaussian entries $\NN(0,1)$ (the $\sqrt n$ scaling ensures $\EE[\| \x \|_2^2] = 1$).
Let $\y \in \RR^n$ be a deterministic vector of unit norm $\| \y \|_2 = 1$. 
We consider two \emph{nonlinear} objects, determined by a nonlinear function $\phi \colon \RR \to \RR$ 
acting in two different regimes:
\begin{enumerate}
  \item 
  \textbf{LLN regime}: 
  here, we consider random variables $f_{\rm LLN}(\x) = \| \x \|_2^2$ or $f_{\rm LLN}(\x) =\x^\top \y$ that both exhibit \emph{LLN-type concentration} (i.e., they are nearly deterministic for $n$ large), and we are interested in the nonlinear $\phi(f_{\rm LLN}(\x))$.
  \item
  \textbf{CLT regime}: 
  here, we consider random variables $f_{\rm CLT}(\x) = \sqrt n(\| \x \|_2^2 - 1)$ or $f_{\rm CLT}(\x) =\sqrt n \cdot \x^\top \y$ that both exhibit \emph{CLT-type concentration} (i.e., they remain inherently random and have \emph{non-degenerate} distributions for $n$ large), and we are interested in the nonlinear $\phi(f_{\rm CLT}(\x))$.
\end{enumerate}
\end{Example}

In the LLN regime, the random variable $f_{\rm LLN}(\x)$ becomes nearly deterministic, and we can  linearize the nonlinear function $\phi(f_{\rm LLN}(\x))$ using Taylor's theorem.\footnote{Strictly speaking, applying Taylor's theorem in \Cref{theo:Taylor} to a random variable $x$ (such as those in \Cref{example:LLN_CLT}) requires a careful argument to ensure that the error term $h_k(x) (x- \tau)^k$ remains $o(|x - \tau|^k)$ with high probability.
This can be justified using concentration arguments.
}

\begin{tcolorbox}[breakable]
\begin{Theorem}[\textbf{Taylor's theorem}]
\label{theo:Taylor}
Let $\phi \colon \RR \to \RR$ be a function that is at least $k$ times continuously differentiable in a neighborhood of some point $\tau \in \RR$. 
Then, there exists $h_k \colon \RR \to \RR$ such that
$\phi(x) = \phi(\tau) + \phi'(\tau) (x - \tau) + \frac{\phi''(\tau)}2  (x - \tau)^2 + \ldots + \frac{\phi^{(k)}(\tau)}{k!} (x - \tau)^k + h_k(x) (x- \tau)^k$ with $\lim_{x \to \tau} h_k(x) = 0$. 
Consequently, $h_k(x) (x- \tau)^k= o(|x - \tau|^k)$ as $x \to \tau$.
\end{Theorem}
\end{tcolorbox}

In the CLT regime, the random variable $f_{\rm CLT}(\x)$ remains random and follows a non-degenerate distribution $f_{\rm CLT}(\x)\sim \mu$.
In this case, we can express $\EE[\phi(\xi)]$ for $ \xi \sim \mu$ as $\EE_{\xi \sim \mu}[\phi(\xi)] = \int \phi(t) \mu(dt)$, and we can linearize $\EE[\phi(\xi)]$ using orthogonal polynomials that are associated with $\mu$.

\begin{tcolorbox}[breakable]
\begin{definition}[\textbf{Orthogonal polynomials and orthogonal polynomial expansion}]
\label{def:ortho_poly}
Let $\mu$ be a probability measure. 
For two functions $\phi$ and $\psi$, define the inner product
\begin{equation}\label{eq:inner_prod_functional}
  \langle \phi,\psi \rangle_\mu \equiv \int \phi(t) \psi(t) \mu(d t) = \EE_{\xi \sim \mu} [ \phi(\xi) \psi(\xi) ].
\end{equation}
Let $\{ P_i( t), i \geq 0 \}$ be a family of \emph{orthogonal polynomials} with respect to this inner product, obtained from the Gram-Schmidt procedure on the monomials \(\{1,t,t^2,\ldots\}\), with \(P_0(t)=1\), and $P_i$ is a polynomial of degree $i$ satisfying $\langle P_i, P_j \rangle = \EE[ P_i(\xi) P_j(\xi) ] = \delta_{ij}$.
Then, for any square-integrable $\phi \in L^2(\mu)$, the \emph{orthogonal polynomial expansion} of $\phi$ is given by
\begin{equation}
\label{eq:def-expansion}
    \phi(\xi) \sim \sum_{i=0}^\infty a_{\phi;i} P_i(\xi), \quad a_{\phi;i} = \int \phi(t) P_i(t) \mu(dt).
\end{equation}
\end{definition}
\end{tcolorbox}
For a standard Gaussian random variable $\xi \sim \NN(0,1)$ with $\mu(dt) = \frac1{\sqrt{2 \pi}} e^{-\frac{t^2}2} dt$, the corresponding orthogonal polynomials are the Hermite polynomials.

\begin{tcolorbox}[breakable]
\begin{Theorem}[\textbf{Hermite polynomial expansion},~\cite{andrews1999special}]
\label{theo:normalized_Hermite}
For $t \in \RR$, the $i^{th}$ normalized Hermite polynomial, denoted ${\rm He}_i(t)$, is given by
\begin{equation}\label{eq:def_Hermite_poly}
  {\rm He}_0(t) = 1, \quad {\rm He}_i(t) = \frac{ (-1)^i }{\sqrt{i!}} e^{\frac{t^2}2 } \frac{d^i}{d t^i} \left( e^{-\frac{t^2}2 } \right), \quad i \geq 1.
\end{equation}
The normalized Hermite polynomials
\begin{enumerate}
  \item are orthogonal polynomials with respect to the standard Gaussian measure, i.e., $\int {\rm He}_i(t) {\rm He}_j(t) \mu(dt) = \delta_{ij}$ for $\mu(dt) = \frac1{\sqrt{2 \pi}} e^{-\frac{t^2}2} dt$; and
  \item can be used to formally expand any square-integrable function $\phi \in L^2(\mu)$ as
  \begin{equation}\label{eq:Hermite_expansion}
    \phi(\xi) \sim \sum_{i = 0}^\infty a_{\phi;i} {\rm He}_i(\xi), \quad a_{\phi;i} = \int \phi(t) {\rm He}_i(t) \mu(dt) = \EE[\phi(\xi) {\rm He}_i(\xi)],
  \end{equation}
  for a standard Gaussian random variable $\xi \sim \NN(0,1)$.
  The coefficients $a_{\phi;i}$ are the \emph{Hermite coefficients} of $\phi$. 
  In particular, $a_{\phi;0} = \EE[\phi(\xi)]$, $a_{\phi;1} = \EE[\xi \phi(\xi)]$, $\sqrt 2 a_{\phi;2} = \EE[\xi^2 \phi(\xi)] - a_{\phi;0}$ and $\nu_\phi = \EE[\phi^2(\xi)] = \sum_{i=0}^{\infty} a_{\phi;i}^2$.
\end{enumerate}
\end{Theorem}
\end{tcolorbox}

\begin{Example}[\textbf{Distinct linearizations of $\tanh$ function in two scaling regimes}]
\label{example:tanh_two_scaling}
As a concrete illustration of linearizing a nonlinear object in the two different scaling regimes in \Cref{def:two_scaling_regimes}, consider $\phi(t) = \tanh(t)$.
By Theorems~\ref{theo:Taylor}~and~\ref{theo:normalized_Hermite}, this nonlinear function is ``close'' to \emph{different} linear functions, depending on the scaling regime.
Concretely, let $\x \in \RR^n$ be a random vector such that $\sqrt n\x$ has i.i.d.\@ standard Gaussian entries, and let $\y \in \RR^n$ be a deterministic vector of unit norm. 
Then:
\begin{enumerate}
  \item
  \textbf{In the LLN regime}, we have for $f_{\rm LLN}(\x) = \x^\top \y$ that $\tanh(f_{\rm LLN}(\x)) - \psi_{\rm LLN}(f_{\rm LLN}(\x)) \to 0$ as $n \to \infty$, with $\psi_{\rm LLN}(t) = t$.
  This is  a consequence of $\tanh(t) \approx t = \psi_{\rm LLN}(t) $ for $t$ close to zero by Taylor expansion in \Cref{theo:Taylor}.
  We also have $\EE[\tanh(f_{\rm LLN}(\x))] - \EE[\psi_{\rm LLN}(f_{\rm LLN}(\x))] \to 0$ as a result.
  \item
  \textbf{In the CLT regime}, we have for $f_{\rm CLT}(\x) = \sqrt n \cdot \x^\top \y$ that $\EE[\tanh(f_{\rm CLT}(\x))] = \EE[\psi_{\rm CLT}(f_{\rm CLT}(\x))]$ and $\EE[\tanh(f_{\rm CLT}(\x)) \cdot f_{\rm CLT}(\x)] = \EE[\psi_{\rm CLT}(f_{\rm CLT}(\x)) \cdot f_{\rm CLT}(\x)]$ in expectation, where the corresponding linear function is $\psi_{\rm CLT}(t) = 0.606 t$.
  Since the two functions share the same zeroth~and~first-order Hermite coefficients, $a_{\tanh;0} = a_{\psi_{\rm CLT};0} = 0$ and $a_{\tanh;1} = a_{\psi_{\rm CLT};1} \approx 0.606$.
\end{enumerate}
\Cref{fig:Taylor_orth_poly} illustrates how $\tanh(f_{\rm LLN}(\x))$ and $\tanh(f_{\rm CLT}(\x))$ compare with their respective linearizations in the LLN and CLT regimes in \Cref{fig:tanh_LLN} and \Cref{fig:tanh_CLT}.
A clear difference can be observed between the two linearizations across the two scaling regimes.
Note that these (high-dimensional) approximations are not unique, since one can further expand (using Theorems~\ref{theo:Taylor}~and~\ref{theo:normalized_Hermite}) to get higher-order, often nonlinear,~terms.
\end{Example}

\begin{figure}[!tb] 
\centering
\begin{subfigure}[c]{0.48\linewidth}
\centering
\begin{tikzpicture}[font=\footnotesize]
    \pgfplotsset{every major grid/.style={style=densely dashed}}
    \begin{axis}[
      axis y line*=left,
      height=.65\linewidth,
      width=.95\linewidth,
      xmin=-3.2,xmax=2.5,
      ymax=2.5,
      ymin=-1.1,
      xtick={-2,0,2},
      bar width=1.5pt,
      ymajorgrids=false,
      grid=major,
      ylabel= \empty,
      legend style = {at={(0.02,0.98)}, anchor=north west, font=\scriptsize}
      ]
      \addplot+[ybar,mark=none,draw=white,fill=black!50!white,area legend] coordinates{
      (-0.105679, 0.615516)(-0.040693, 5.796109)(0.024293, 7.745243)(0.089279, 1.179739)(0.154265, 0.051293)};
      \addlegendentry{{ $f_{\rm LLN}(\x) \equiv \x^\top \y$ }};
      \addplot[smooth,samples=300,domain=-2.5:2.5,line width=1.5pt,BLUE] { tanh(x) };
      \addlegendentry{{ $\phi(t) = \tanh(t)$ }};
      \addplot[smooth,samples=300,domain=-2.5:2.5,line width=1.5pt,RED] {  x };
      \addlegendentry{{ $\psi_{\rm LLN}(t) = t$ }};
    \end{axis}
  \end{tikzpicture}
  \caption{ LLN regime }
  \label{fig:tanh_LLN}
  \end{subfigure}
  \hfill
  \begin{subfigure}[c]{0.48\linewidth}
  \centering
  \begin{tikzpicture}[font=\footnotesize]
    \pgfplotsset{every major grid/.style={style=densely dashed}}
    \begin{axis}[
      axis y line*=left,
      height=.65\linewidth,
      width=.95\linewidth,
      xmin=-3.2,xmax=2.5,
      ymax=2.5,
      ymin=-1.1,
      xtick={-2,0,2},
      bar width=2.5pt,
      ymajorgrids=false,
      grid=major,
      ylabel= \empty,
      legend style = {at={(0.02,0.98)}, anchor=north west, font=\scriptsize}
      ]
      \addplot+[ybar,mark=none,draw=white,fill=black!50!white,area legend] coordinates{
      (-2.589377, 0.066858)(-2.489664, 0.000000)(-2.389950, 0.033429)(-2.290237, 0.000000)(-2.190523, 0.066858)(-2.090810, 0.000000)(-1.991097, 0.066858)(-1.891383, 0.033429)(-1.791670, 0.066858)(-1.691956, 0.100287)(-1.592243, 0.066858)(-1.492530, 0.066858)(-1.392816, 0.000000)(-1.293103, 0.200575)(-1.193390, 0.167146)(-1.093676, 0.334291)(-0.993963, 0.200575)(-0.894249, 0.267433)(-0.794536, 0.334291)(-0.694823, 0.234004)(-0.595109, 0.401150)(-0.495396, 0.468008)(-0.395682, 0.401150)(-0.295969, 0.234004)(-0.196256, 0.501437)(-0.096542, 0.434579)(0.003171, 0.367721)(0.102885, 0.200575)(0.202598, 0.468008)(0.302311, 0.534866)(0.402025, 0.534866)(0.501738, 0.401150)(0.601451, 0.367721)(0.701165, 0.267433)(0.800878, 0.334291)(0.900592, 0.300862)(1.000305, 0.300862)(1.100018, 0.200575)(1.199732, 0.133717)(1.299445, 0.267433)(1.399159, 0.200575)(1.498872, 0.033429)(1.598585, 0.066858)(1.698299, 0.000000)(1.798012, 0.100287)(1.897725, 0.033429)(1.997439, 0.033429)(2.097152, 0.033429)(2.196866, 0.066858)(2.296579, 0.033429)};
      \addlegendentry{{ $f_{\rm CLT}(\x) \equiv \sqrt n \cdot \x^\top \y$ }};
      \addplot[smooth,samples=300,domain=-2.5:2.5,line width=1.5pt,BLUE] {  tanh(x) };
      \addlegendentry{{ $\phi(t) = \tanh(t)$ }};
      \addplot[smooth,samples=300,domain=-2.5:2.5,line width=1.5pt,RED] {  x*0.606 };
      \addlegendentry{{ $\psi_{\rm CLT}(t) = 0.606 t$ }};
    \end{axis}
  \end{tikzpicture}
\caption{ CLT regime }
\label{fig:tanh_CLT}
\end{subfigure}
\caption{Different behavior of nonlinear $\phi(f_{\rm LLN}(\x))$ and $\phi(f_{\rm CLT}(\x))$ for $\phi(t) = \tanh(t)$ (in \textbf{\BLUE blue}) in the LLN and CLT regime, with $n = 500$. 
We have $\tanh(f_{\rm LLN}(\x)) \simeq \psi_{\rm LLN}(f_{\rm LLN}(\x))$ in the LLN regime by Taylor expansion, and $\EE[\tanh(f_{\rm CLT}(\x))] = \EE[\psi_{\rm CLT}(f_{\rm CLT}(\x))], \EE[\tanh(f_{\rm CLT}(\x)) \cdot f_{\rm CLT}(\x)] = \EE[\psi_{\rm CLT}(f_{\rm CLT}(\x)) \cdot f_{\rm CLT}(\x)]$ in the CLT regime (as a consequence of $a_{\tanh;0} = a_{\psi_{\rm CLT};0}$ and $a_{\tanh;1} = a_{\psi_{\rm CLT};1}$), with \emph{different} linear functions $\psi_{\rm LLN}(t) = t$ and $\psi_{\rm CLT}(t) = 0.606 t$ in \textbf{\RED red}.
}
\label{fig:Taylor_orth_poly}
\end{figure}

Like the Deterministic Equivalent in \Cref{def:DE}, the High-dimensional Linearizations presented in \Cref{example:tanh_two_scaling} and visualized in \Cref{fig:Taylor_orth_poly} are special cases of the High-Dimensional Equivalent in \Cref{def:HiE}.

\begin{tcolorbox}[breakable]
\begin{definition}[\textbf{Linear Equivalent}]\label{def:Hi-LE}
A \emph{Linear Equivalent} is a special case of the \emph{High-Dimensional Equivalent} in \Cref{def:HiE}, for two nonlinear random matrix models $\mathcal{M}_{\phi}(\X)$ and $\tilde{\mathcal{M}}_{\psi}(\X)$, where $\psi$ may be different from $\phi$ so that $\tilde{\mathcal{M}}_{\psi}(\X)$ becomes more amenable to analysis.
We denote 
\begin{equation*}
  f\left( \mathcal{M}_{\phi}(\X) \right) - f ( \tilde{\mathcal{M}}_{\psi}(\X) ) \to 0~\text{as}~n,p\to\infty\quad  \Leftrightarrow \quad \mathcal{M}_{\phi}(\X) \overset{f}{\leftrightarrow} \tilde{\mathcal{M}}_{\psi}(\X) ~\text{or}~ \mathcal{M}_{\phi}(\X)\leftrightarrow \tilde{\mathcal{M}}_{\psi}(\X).
\end{equation*}
\end{definition}
\end{tcolorbox}
As already mentioned in \Cref{example:tanh_two_scaling}, for a given nonlinear random matrix model, its linear equivalent is \emph{not} unique, and it may \emph{not} even be a ``linear'' functions despite the name.

\ifisarxiv
\section{Linear Random Matrix Model: A Deterministic Equivalent Approach}\label{sec:linear_model}
\fi

In this section, we present Deterministic Equivalent results for linear random matrix models, and we illustrate how these results apply to the analysis of linear ML models in the proportional regime.

\ifisarxiv
\subsection{Deterministic Equivalent for resolvents of linear sample covariance and Gram matrices}
\fi

Consider a matrix $\X = [\x_1, \ldots, \x_n] \in \RR^{p \times n}$ composed of $n$ random vectors of dimension $p$. 
We define the sample covariance matrix (SCM) $\hat \C \in \RR^{p \times p}$, the Gram matrix $\G \in \RR^{n \times n}$, and their resolvents as
\begin{align}
  \hat \C &\equiv \frac1n \X \X^\top \in \RR^{p \times p}, \quad \Q_{\hat \C} \equiv (\hat \C - z \I_p )^{-1}, \label{eq:def_SCM} \\ 
  \G &\equiv \frac1n \X^\top \X \in \RR^{n \times n}, \quad \Q_{\G} \equiv (\G - z \I_n )^{-1}. \label{eq:def_Gram}
\end{align}
We have the following Deterministic Equivalent results.
\ifisarxiv
See \Cref{sec:proof_G_SCM_DE} for details of the proof.
\fi

\begin{tcolorbox}[breakable]
\begin{Theorem}[\textbf{Deterministic Equivalents for SCM and Gram resolvents}]
\label{theo:G_SCM_DE}
Let $\Z \in \RR^{p \times n}$ be a random matrix with independent sub-gaussian\footnotemark~entries having zero mean and unit variance, and consider a deterministic, positive-definite matrix $\C \in \RR^{p \times p}$ of bounded spectral norm. 
Define $\X =\C^{\frac12} \Z$.
Then for $z \in \CC \setminus (0,\infty)$ as $n,p \to \infty$ with $p/n \to c \in (0,\infty)$, the resolvents $\Q_{\hat \C}(z), \Q_{\G}(z)$ of the SCM $\hat \C$ and the Gram matrix $\G$ admit the following Deterministic Equivalents: 
\begin{align}
  \Q_{\hat \C}(z) &\overset{f}{\leftrightarrow} \tilde \Q_{\hat \C}(z), \quad \tilde \Q_{\hat \C}(z) = \left( \frac{\C}{1+\delta(z)} - z \I_p \right)^{-1}, \label{eq:def_bar_Q_SCM} \\ 
  \Q_{\G}(z) &\overset{f}{\leftrightarrow} \tilde \Q_{\G}(z), \quad \tilde \Q_{\G}(z) = -\frac{\I_n}{z (1 + \delta(z))} , \label{eq:def_bar_Q_G}
\end{align}
for trace and bilinear forms $f(\Q) = \tr(\A \Q)/n, f(\Q) = \mathbf{a}^\top \Q \mathbf{b}$ of $\Q$, with $\A, \mathbf{a}, \mathbf{b}$ of bounded spectral and Euclidean norms as in \Cref{rem:scalar_func}, where $\delta(z)$ is the unique solution of $\delta(z) = \frac1n \tr \left(\tilde \Q_{\hat \C}(z) \C \right)$ corresponding to the Stieltjes transform of a probability measure (see \Cref{rem:ST}).
\end{Theorem}
\end{tcolorbox}
\footnotetext{We say $z$ is a \emph{sub-gaussian random variable} if it has a tail that decays \emph{as fast as} standard Gaussian random variables, that is $\Pr \left( |z| \geq t \right) \leq 2 \exp(-t^2/\sigma_{\NN}^2)$ for all $t > 0$, and the \emph{sub-gaussian norm} of $z$ is the \emph{smallest} $\sigma_{\NN} > 0$ such that this holds.}
\begin{Remark}[\textbf{Special case: $\C = \I_p$ and the Mar{\u c}enko-Pastur law}]\label{rem:MP}
\normalfont
When $\C = \I_p$, \Cref{theo:G_SCM_DE} implies that $\delta(z) = c m(z)$ and $\tilde \Q_{\hat \C}(z) = m(z) \I_p$, where $m(z)$ is the unique Stieltjes transform solution to
\begin{equation}\label{eq:MP}
  c z m^2(z) - (1 - c - z) m(z) + 1 =0 .
\end{equation}
This is the Mar{\u c}enko-Pastur (MP) equation~\cite{marvcenko1967distribution}, and $m(z)$ is the Stieltjes transform of the MP law 
\begin{equation}\label{eq:MP-law}
    \mu(dx) = (1-c^{-1})^+ \delta_0(x) + \frac1{2\pi c x} \sqrt{ \left(x-  E_- \right)^+ \left( E_+ -x \right)^+}\,dx,
\end{equation}
where $E_\pm =  (1 \pm \sqrt c)^2$ and $(x)^+ = \max(0,x)$. 
\Cref{fig:MP-law-different-c} shows this distribution for different $c = \lim p/n$.
\end{Remark}

\begin{figure}[!tb]
\centering
\begin{tikzpicture}[font=\footnotesize]
\renewcommand{\axisdefaulttryminticks}{4} 
\pgfplotsset{every major grid/.style={densely dashed}}       
\tikzstyle{every axis y label}+=[yshift=-10pt] 
\tikzstyle{every axis x label}+=[yshift=5pt]
\pgfplotsset{every axis legend/.style={cells={anchor=west},fill=white,
at={(0.98,0.98)}, anchor=north east, font=\footnotesize}}
\begin{axis}[
width=.6\linewidth,
height=.3\linewidth,
xmin=0,
ymin=0,
xmax=6,
ymax=1.2,
grid=major,
ymajorgrids=false,
scaled ticks=true,
xlabel={$x$},
ylabel={$\mu(dx)$}
]
\def\c{.1}
\addplot[samples=300,domain=0:6,line width=1pt] {1/(2*pi*\c*x)*sqrt(max(((1+sqrt(\c))^2-x)*(x-(1-sqrt(\c))^2),0))};
\def\c{.5}
\addplot[densely dashed,samples=300,domain=0:6,line width=1pt] {1/(2*pi*\c*x)*sqrt(max(((1+sqrt(\c))^2-x)*(x-(1-sqrt(\c))^2),0))};
\def\c{1}
\addplot[densely dotted,samples=300,domain=0.01:6,line width=1pt] {1/(2*pi*\c*x)*sqrt(max(((1+sqrt(\c))^2-x)*(x-(1-sqrt(\c))^2),0))};
\def\c{2}
\addplot[semitransparent,samples=300,domain=0:6,line width=1pt] {1/(2*pi*\c*x)*sqrt(max(((1+sqrt(\c))^2-x)*(x-(1-sqrt(\c))^2),0))};
\legend{ {$c=0.1$},{$c=0.5$},{$c=1$},{$c=2$} }
\end{axis}
\end{tikzpicture}
\caption{ {Mar{\u{c}}enko-Pastur law in \Cref{eq:MP-law} for different values of $c$.} }
\label{fig:MP-law-different-c}
\end{figure}

\begin{Remark}[\textbf{SCM in the classical versus proportional regime}]\label{rem:SCM_classical_proportional}
\normalfont
Let $\C = \I_p$. 
We then observe the following contrasting behaviors for the SCM $\hat \C = \frac1n \X \X^\top$ in \eqref{eq:def_SCM}, depending on the ratio $p/n$:
\begin{enumerate}
  \item in the \emph{classical regime} with $p$ fixed as $n \to \infty$, we have, by the LLN that $\hat \C \to \EE[\hat \C] = \I_p$, so that the resolvent $\Q_{\hat \C} \to (1 - z)^{-1} \I_p$ and is approximately deterministic in this regime; and
  \item in the \emph{proportional regime} as $n,p \to \infty$ with $p/n \to c  \in (0, \infty)$, we have that $\hat \C$ does \emph{not} converge to $\EE[\hat \C]$ in a spectral norm sense, and $\Q_{\hat \C}$ is no longer deterministic.
  Nonetheless, it admits a \emph{Deterministic Equivalent} (in the sense of \Cref{def:DE}) $\hat \Q_{\hat \C}$ given in \Cref{theo:G_SCM_DE}, parameterized by $c = \lim p/n$.
  Notably, taking $c \to 0$ recovers the classical-regime result as a special case.
\end{enumerate}
\end{Remark}

\begin{Remark}[\textbf{High-dimensional universality}]\label{rem:universality}
\normalfont
In \Cref{theo:G_SCM_DE}, we do not assume any specific distribution for the random matrix $\Z$, other than requiring independent sub-gaussian entries of zero mean and unit variance.
Accordingly, the Deterministic Equivalents in \Cref{theo:G_SCM_DE}—and the resulting eigenspectral functionals from \Cref{theo:spectral_func_contour}—depend on the distribution of $\Z$ \emph{only through its first-~and~second-order moments}.
This \emph{universal} property is well-known in RMT and high-dimensional statistics~\cite{tao2010random,pastur2011eigenvalue,bai2010spectral}. 
Moreover, there is a growing interest in the \emph{high-dimensional universality} of ML models ranging from generalized linear models to shallow and deep neural networks, as reflected in recent work~\cite{goldt2022gaussian,hu2022universality,montanari2022universality,mai2025the}.
\end{Remark}

\ifisarxiv
\subsection{Linear least squares: classical versus proportional regime}
\fi

Linear least squares regression—also known as linear ridge regression when an $\ell_2$ regularization term is added—is among the most widely used methods in applied mathematics, statistics, data analysis, and ML. 
Despite its extensive applications, the statistical behavior of linear (least squares) regression in the \emph{proportional regime}, with $n \sim p$ both large, has only recently begun to receive close attention~\cite{dobriban2018high,hastie2022Surprises}.

Below, we provide a detailed characterization of both in-sample and out-of-sample prediction risks for linear regression, contrasting the classical regime ($n\gg p$) with the proportional regime ($n \sim p$).

\begin{tcolorbox}[breakable]
\begin{definition}[\textbf{Homogeneous noisy linear model}]
\label{def:noisy_linear_model}
Let $\{ (\x_i, y_i) \}_{i=1}^n$ be $n$ data-target pairs, where each $\x_i \in \RR^p$ has independent sub-gaussian entries of zero mean and unit variance, and $y_i$ is generated from the following noisy linear model:
\begin{equation}\label{eq:linear_model}
  y = \bbeta_*^\top \x + \epsilon ,
\end{equation}
for some deterministic ground-truth vector $\bbeta_* \in \RR^p$. 
The noise $\epsilon \in \RR$ is independent of $\x \in \RR^p$, with $\EE[\epsilon] = 0$ and $\Var[\epsilon]= \sigma^2$.
Given any $\bbeta \in \RR^p$, we define:
\begin{enumerate}
  \item \textbf{in-sample prediction risk}: $R_{\rm in}(\bbeta) = \frac1n \| \bbeta^\top \X - \bbeta_*^\top \X \|_2^2$; and
  \item \textbf{out-of-sample prediction risk}: $R_{\rm out}(\bbeta) = \EE[(\bbeta^\top \x' - \bbeta_*^\top \x')^2~|~\X]$;
\end{enumerate}
where $\X = [\x_1, \ldots, \x_n] \in \RR^{p \times n}$ is the training data, and $\x' \in \RR^p$ is an \emph{independent} test sample. 
\end{definition}
\end{tcolorbox}

\noindent 
We consider the linear regressor that minimizes the following ridge-regularized MSE
\begin{equation}\label{eq:linear_reg}
  L(\bbeta) = \frac1n \sum_{i=1}^n (y_i - \bbeta^\top \x_i)^2 + \gamma \| \bbeta \|_2^2 = \frac1n \| \y - \X^\top \bbeta \|_2^2 + \gamma \| \bbeta \|_2^2 ,
\end{equation}
for $\y = [y_1, \ldots, y_n]^\top \in \RR^n$ and $\gamma \geq 0$, the regularization parameter (so $\gamma = 0$ corresponds to ``ridgeless'' least squares regression). 
The solution is uniquely given by
\begin{equation}\label{eq:def-linear_reg_beta}
  \bbeta_\gamma = \left(\frac1n \X \X^\top + \gamma \I_p \right)^{-1} \frac1n \X \y = \frac1n \X \left(\frac1n \X^\top \X + \gamma \I_n \right)^{-1} \y ,
\end{equation}
assuming both inverses exist (which is guaranteed for $\gamma > 0$). 
In the case $\gamma = 0$, we consider the \emph{minimum $\ell_2$-norm} least squares solution:
\begin{equation}\label{eq:def-ridgeless_beta}
  \bbeta_0 = \left( \X \X^\top \right)^+ \X \y =  \X \left( \X^\top \X \right)^+ \y,
\end{equation}
where $(\cdot)^+$ denotes the Moore–Penrose pseudoinverse.
This ``ridgeless'' least squares solution can also be obtained by taking the limit $\gamma \to 0$ in \eqref{eq:def-linear_reg_beta}.

The following result characterizes the asymptotic behavior of the in-sample and out-of-sample risks of the linear regressor $\bbeta_\gamma$ in \eqref{eq:def-linear_reg_beta}, under the homogeneous noisy linear model in \Cref{def:noisy_linear_model}.
We consider here both the \emph{classical regime} ($n \gg p$) and the \emph{proportional regime} ($n \sim p$, where $n$ could be greater than or less than $p$).
\ifisarxiv
In particular, the proof follows from the Deterministic Equivalent result in \Cref{theo:G_SCM_DE} with $f(\cdot)$ taken to be either the normalized trace or a bilinear form.
See \Cref{sec:proof_of_risk_LS} for a detailed derivation.
\fi 

\begin{tcolorbox}[breakable]
\begin{Proposition}[\textbf{Risk of linear ridge regression}]
\label{prop:risk_LS}
Under the homogeneous noisy linear model in \Cref{def:noisy_linear_model}, let $R_{\rm in}(\bbeta_\gamma)$ and $R_{\rm out}(\bbeta_\gamma)$ be the in-sample and out-of-sample risks, respectively, for the linear regressor $\bbeta_\gamma$ defined in \eqref{eq:def-linear_reg_beta}. 
Then,
\begin{enumerate}
  \item 
  in the \textbf{classical} regime, for $p$ fixed and as $n \to \infty$, it holds that $R_{\rm in}(\bbeta_\gamma) - R_{{\rm in},n \gg p}(\gamma) \to 0$ and $R_{\rm out}(\bbeta_\gamma) - R_{{\rm out},n \gg p}(\gamma) \to 0$ almost surely, where
  \begin{equation*}
    R_{{\rm in}, n \gg p}(\gamma) = R_{{\rm out}, n \gg p}(\gamma) = \frac{ \gamma^2 \| \bbeta_*\|_2^2 +  \frac{p}n \sigma^2 }{(1+\gamma)^2};
  \end{equation*}
  \item 
  in the \textbf{proportional} regime, as $n,p \to \infty$ with $p/n \to c \in (0,1) \cup (1, \infty)$, it holds that $R_{\rm in}(\bbeta_\gamma) - R_{{\rm in},n \sim p}(\gamma) \to 0$ and $R_{\rm out}(\bbeta_\gamma) - R_{{\rm out},n \sim p}(\gamma) \to 0$ almost surely, where
   \begin{align*}
    R_{{\rm in}, n \sim p}(\gamma) &= \sigma^2 c + \gamma m(-\gamma) (\gamma \| \bbeta_* \|_2^2 - 2 \sigma^2 c ) + \gamma^2 m'(-\gamma ) ( \gamma \| \bbeta_* \|_2^2 - \sigma^2 c )  , \\ 
    R_{{\rm out}, n \sim p}(\gamma) &= \sigma^2 c m(-\gamma) + \gamma m'(-\gamma) \left( \sigma^2 c - \gamma \| \bbeta_* \|_2^2 \right).
  \end{align*}
  Here, $m'(-\gamma) = \frac{m(-\gamma) (c m(-\gamma) + 1 )}{ 2 c \gamma m(-\gamma) + 1 - c + \gamma }$ is obtained by differentiating the MP equation in \eqref{eq:MP}.
\end{enumerate}
\end{Proposition}
\end{tcolorbox}
From \Cref{prop:risk_LS}, we see that the asymptotic prediction risks behave differently in the classical ($n \gg p$) versus the proportional ($n \sim p$) regime.
Classical asymptotic statistics can predict the former scenario, whereas the RMT analysis framework is able to capture the latter scenario.

\begin{Remark}[\textbf{Scaling law of in-sample risk}]\label{rem:scaling_in_sample}
\normalfont
Taking $\gamma = 0$ in the classical and proportional regimes in \Cref{prop:risk_LS}, we find for $n > p$ that $R_{{\rm in}, n \gg p}(\gamma = 0) = R_{{\rm in}, n \sim p}(\gamma = 0) = \sigma^2 \frac{p}n$.
Thus, for fixed $\sigma^2 p$, the in-sample risk decays as $n^{-1}$ as $n$ grows large.
This scaling law does not hold for $\gamma \gg 0$. 
\ifisarxiv
See \Cref{subfig:in_sample_risk} for a numerical illustration and \Cref{sec:proof_of_risk_LS} for a detailed derivation.
\fi
\end{Remark}

\begin{Remark}[\textbf{Double descent out-of-sample risk in the proportional regime}]\label{rem:double_descent_out-of-sample}
\normalfont
Consider now the out-of-sample risk $R_{\rm out}$.
From \Cref{prop:risk_LS}, for $n \sim p$ both large, we have $R_{{\rm out}, n \sim p}(\gamma \to 0) = \| \bbeta_* \|_2^2 \max\left(1- c^{-1},0 \right) + \sigma^2 \frac1{\max(c,c^{-1}) - 1}$. 
Hence, the out-of-sample risk of ``ridgeless'' linear regression exhibits a \emph{double-descent} curve as a function of the ratio $n/p$, with a singularity at $c = p/n = 1$.
In other words, in the modern proportional regime, increasing the sample size $n$ can paradoxically \emph{worsen} the model's generalization performance near $n=p$.
This starkly contrasts with the classical regime, where the out-of-sample risk \emph{monotonically decreases} with $n$.
This unexpected phenomenon, known as ``double descent''~\cite{belkin2019reconciling,hastie2022Surprises,liao2020random}, has garnered significant attention in the ML community.
\Cref{subfig:out_of_sample_risk} illustrates that for $n,p$ both large and a modest regularization $\gamma = 10^{-5}$, the out-of-sample risk:
\ifisarxiv
\begin{enumerate}
  \item decreases, then increases as $n$ approaches $p$ in the \emph{under-determined} $n<p$ regime; 
  \item reaches a singular ``peak point'' at $n=p$ with a large risk; 
  \item decreases again in the \emph{over-determined} $n > p$ regime. 
\end{enumerate}
For larger regularization ($\gamma = 10^{-1}$), this effect is significantly mitigated.
\fi
\end{Remark}
\begin{figure}[!tb]
\centering
\begin{subfigure}[t]{0.48\linewidth}
\centering
\begin{tikzpicture}[font=\footnotesize]
    \pgfplotsset{every major grid/.style={style=densely dashed}}
    \begin{axis}[
      height=.7\linewidth,
      width=.95\linewidth,
      xmin=0.2,
      xmax=20,
      ymin=0.004,
      ymax=0.11,
      xmode=log,
      ymode=log,
      grid=major,
      scaled ticks=true,
      xlabel={ $n/p$ },
      ylabel={ In-sample risk },
      legend style = {at={(0,0)}, anchor = south west, font=\footnotesize},
      ]
      \addplot[BLUE,mark=o,only marks,line width=1pt] coordinates{
      (0.195312,0.101339)(0.228516,0.098439)(0.267578,0.103521)(0.314453,0.099296)(0.367188,0.098265)(0.431641,0.098915)(0.505859,0.099347)(0.591797,0.098441)(0.695312,0.100059)(0.814453,0.099978)(0.955078,0.102368)(1.119141,0.089263)(1.312500,0.075785)(1.539062,0.065359)(1.802734,0.055065)(2.113281,0.047653)(2.476562,0.039982)(2.904297,0.034727)(3.404297,0.029331)(3.990234,0.025285)(4.677734,0.020840)(5.482422,0.018268)(6.425781,0.015866)(7.531250,0.013213)(8.828125,0.011202)(10.347656,0.009524)(12.128906,0.008096)(14.214844,0.007003)(16.662109,0.006056)(19.531250,0.005031)
      };
      \addlegendentry{{ Empirical $\gamma = 10^{-5}$ }};
      \addplot[BLUE,densely dashed,line width=1pt] coordinates{
      (0.195312,0.100016)(0.228516,0.100015)(0.267578,0.100014)(0.314453,0.100013)(0.367188,0.100011)(0.431641,0.100010)(0.505859,0.100008)(0.591797,0.100005)(0.695312,0.100002)(0.814453,0.099995)(0.955078,0.099959)(1.119141,0.089338)(1.312500,0.076184)(1.539062,0.064971)(1.802734,0.055469)(2.113281,0.047318)(2.476562,0.040377)(2.904297,0.034431)(3.404297,0.029374)(3.990234,0.025061)(4.677734,0.021377)(5.482422,0.018240)(6.425781,0.015562)(7.531250,0.013278)(8.828125,0.011327)(10.347656,0.009664)(12.128906,0.008245)(14.214844,0.007035)(16.662109,0.006002)(19.531250,0.005120)
      };
      \addlegendentry{{ $R_{{\rm in}, n \sim p}(\gamma = 0)$ }};
      \addplot[RED,mark=triangle,only marks,line width=1pt] coordinates{
      (0.195312,0.094153)(0.228516,0.093441)(0.267578,0.093021)(0.314453,0.093993)(0.367188,0.091875)(0.431641,0.091054)(0.505859,0.088312)(0.591797,0.086224)(0.695312,0.080979)(0.814453,0.076352)(0.955078,0.076281)(1.119141,0.068129)(1.312500,0.063532)(1.539062,0.058174)(1.802734,0.052057)(2.113281,0.045670)(2.476562,0.041049)(2.904297,0.036998)(3.404297,0.033062)(3.990234,0.029701)(4.677734,0.026329)(5.482422,0.023836)(6.425781,0.021427)(7.531250,0.019730)(8.828125,0.018154)(10.347656,0.016595)(12.128906,0.015398)(14.214844,0.014354)(16.662109,0.013469)(19.531250,0.012800)
      };
      \addlegendentry{ { Empirical $\gamma = 10^{-1}$ } };
      \addplot[RED,densely dashed,line width=1pt] coordinates{
      (0.195312,0.095802)(0.228516,0.095025)(0.267578,0.094086)(0.314453,0.092923)(0.367188,0.091567)(0.431641,0.089846)(0.505859,0.087784)(0.591797,0.085312)(0.695312,0.082254)(0.814453,0.078669)(0.955078,0.074368)(1.119141,0.069335)(1.312500,0.063659)(1.539062,0.057717)(1.802734,0.051910)(2.113281,0.046421)(2.476562,0.041423)(2.904297,0.036934)(3.404297,0.032986)(3.990234,0.029532)(4.677734,0.026527)(5.482422,0.023929)(6.425781,0.021687)(7.531250,0.019757)(8.828125,0.018097)(10.347656,0.016673)(12.128906,0.015452)(14.214844,0.014408)(16.662109,0.013513)(19.531250,0.012747)
      };
      \addlegendentry{{ $R_{{\rm in}, n \sim p}(\gamma = 10^{-1})$ }};
    \end{axis}
  \end{tikzpicture}
  \caption{{In-sample risk}} \label{subfig:in_sample_risk}
\end{subfigure}
\hfill{}
\begin{subfigure}[t]{0.48\linewidth}
\centering
\begin{tikzpicture}[font=\footnotesize]
    \pgfplotsset{every major grid/.style={style=densely dashed}}
    \begin{axis}[
      height=.7\linewidth,
      width=.95\linewidth,
      xmin=0,
      xmax=2,
      ymin=0,
      ymax=2,
      grid=major,
      scaled ticks=true,
      xlabel={ $n/p$ },
      ylabel={ Out-of-sample risk },
      legend style = {at={(1.28,1)}, anchor = north east, font=\footnotesize},
      ]
      \addplot[BLUE,mark=o,only marks,line width=1pt] coordinates{
          (0.001953,0.993490)(0.072266,0.944360)(0.144531,0.874375)(0.214844,0.811062)(0.287109,0.752026)(0.357422,0.705636)(0.429688,0.644374)(0.500000,0.588635)(0.572266,0.568700)(0.642578,0.531414)(0.714844,0.546836)(0.785156,0.595778)(0.857422,0.762364)(0.927734,1.292994)(1.000000,16.097398)(1.048828,2.118365)(1.117188,0.885045)(1.185547,0.536954)(1.251953,0.405941)(1.320312,0.326435)(1.388672,0.255892)(1.457031,0.213225)(1.523438,0.188200)(1.591797,0.166669)(1.660156,0.151812)(1.728516,0.137006)(1.794922,0.127380)(1.863281,0.119943)(1.931641,0.104371)(2.000000,0.100885)
      };
      \addlegendentry{{ Empirical $\gamma = 10^{-5}$ }};
      \addplot[BLUE,densely dashed,line width=1pt] coordinates{
      (0.001953,0.998243)(0.072266,0.935524)(0.144531,0.872364)(0.214844,0.812519)(0.287109,0.753165)(0.357422,0.698201)(0.429688,0.645655)(0.500000,0.600000)(0.572266,0.561524)(0.642578,0.537203)(0.714844,0.535841)(0.785156,0.580298)(0.857422,0.743948)(0.927734,1.356049)(1.000000,3)(1.048828,2.048000)(1.117188,0.853333)(1.185547,0.538947)(1.251953,0.396899)(1.320312,0.312195)(1.388672,0.257286)(1.457031,0.218803)(1.523438,0.191045)(1.591797,0.168977)(1.660156,0.151479)(1.728516,0.137265)(1.794922,0.125799)(1.863281,0.115837)(1.931641,0.107338)(2.000000,0.100000)
      };
      \addlegendentry{{ $R_{{\rm out},n \sim p}(\gamma \to 0)$ }};
      \addplot[RED,mark=triangle,only marks,line width=1pt] coordinates{
      (0.001953,0.996895)(0.072266,0.934370)(0.144531,0.866766)(0.214844,0.797522)(0.287109,0.751279)(0.357422,0.690205)(0.429688,0.630217)(0.500000,0.579457)(0.572266,0.524558)(0.642578,0.472635)(0.714844,0.425414)(0.785156,0.390997)(0.857422,0.334669)(0.927734,0.307187)(1.000000,0.271368)(1.048828,0.254693)(1.117188,0.224013)(1.185547,0.208164)(1.251953,0.189177)(1.320312,0.168230)(1.388672,0.159590)(1.457031,0.147039)(1.523438,0.139964)(1.591797,0.128371)(1.660156,0.116668)(1.728516,0.111409)(1.794922,0.107529)(1.863281,0.098365)(1.931641,0.095928)(2.000000,0.093373)
      };
      \addlegendentry{ { Empirical $\gamma = 10^{-1}$ } };
      \addplot[RED,densely dashed,line width=1pt] coordinates{
      (0.001953,0.998242)(0.072266,0.935399)(0.144531,0.871764)(0.214844,0.810915)(0.287109,0.749629)(0.357422,0.691390)(0.429688,0.633157)(0.500000,0.578293)(0.572266,0.524022)(0.642578,0.473617)(0.714844,0.424728)(0.785156,0.380511)(0.857422,0.339026)(0.927734,0.302870)(1.000000,0.270156)(1.048828,0.250538)(1.117188,0.226219)(1.185547,0.205227)(1.251953,0.187622)(1.320312,0.171962)(1.388672,0.158425)(1.457031,0.146676)(1.523438,0.136705)(1.591797,0.127693)(1.660156,0.119754)(1.728516,0.112724)(1.794922,0.106636)(1.863281,0.101022)(1.931641,0.095979)(2.000000,0.091427)
      };
      \addlegendentry{{ $R_{{\rm out}, n \sim p}(\gamma = 10^{-1})$ }};
      \addplot[samples=200,domain=0:2,densely dashed,black,line width=1pt] {0.1/(x)};
      \addlegendentry{{ $R_{{\rm out},n \gg p} (\gamma = 0)$ } };
    \end{axis}
  \end{tikzpicture}
  \caption{{Out-of-sample risk}} \label{subfig:out_of_sample_risk}
  \end{subfigure}
\caption{{
Empirical and theoretical risks of the linear regressor $\bbeta_\gamma$ in \eqref{eq:def-linear_reg_beta} as a function of the ratio $n/p$, for $\gamma = 10^{-1}$ and $\gamma = 10^{-5}$, with $p = 512$, $\| \bbeta_* \|_2 = 1$, Gaussian $\x \sim \mathcal N(\zo, \I_p)$, and $\varepsilon \sim \mathcal N(0,\sigma^2=0.1)$.
\textbf{\Cref{subfig:in_sample_risk}}: log-log plot of in-sample risk averaged over $30$ runs.
\textbf{\Cref{subfig:out_of_sample_risk}}: out-of-sample risk averaged over $30$ runs on independent test sets of size $n' = 2\,048$.
}}
\label{fig:linear-reg}
\end{figure}

\ifisarxiv
\section{Single-hidden-layer NN Model: Deterministic Equivalent and Linearization}\label{sec:shallow_NN}
\fi

In this section, we demonstrate how the proposed random matrix analysis framework can be extended to \emph{nonlinear} single-hidden-layer NN models.
We first use a Deterministic Equivalent approach for the nonlinear resolvent to evaluate the training and test errors.
We then apply the high-dimensional linearization techniques discussed earlier to the resulting nonlinear kernel matrices, examining how nonlinear activations affect the NN performance.

\begin{tcolorbox}[breakable]
\begin{definition}[\textbf{Single-hidden-layer NN model}]\label{def:single-layer-NN}
Consider a single-hidden-layer NN model with first-layer weights $\W \in \RR^{d \times p}$ and second-layer weights $\bbeta \in \RR^d$. 
For an input vector $\x \in \RR^p$, the network output is given by $\hat y(\x) = \bbeta^\top \phi(\W \x)$, where $\phi(\cdot)$ is an entrywise activation function.
We are interested in the NN performance measured by:
\begin{enumerate}
  \item its \textbf{training MSE} $E_{\train} = \frac1n \sum_{i=1}^n (y_i - \hat y(\x_i))^2 = \frac1n \| \y - \bPhi^\top \bbeta \|_2^2$ with $ \bPhi \equiv \phi(\W \X)$ for a training set $(\X, \y)$ of size $n$, $\X = [\x_1, \ldots, \x_n] \in \RR^{p \times n}, \y = [y_1, \ldots, y_n]^\top \in \RR^n$; and 
  \item its \textbf{test MSE} $E_{\test}  = \frac1{n'} \sum_{i=1}^{n'} (y_i' - \hat y(\x_i'))^2 = \frac1{n'} \| \y' - \phi(\W \X')^\top \bbeta \|_2^2$ on a test set $(\X', \y')$ of size $n'$, with $\X' = [\x_1', \ldots, \x_{n'}'] \in \RR^{p \times n'} $ and $ \y' = [y_1', \ldots, y_{n'}']^\top \in \RR^{n'}$.
\end{enumerate}
\end{definition}
\end{tcolorbox}
Unlike in \Cref{def:noisy_linear_model} for the linear model, here we condition on $(\X,\y)$ and treat the training data and targets $\X \in \RR^{p \times n}, \y \in \RR^n$ as \emph{deterministic}, and consider that the network weight matrix $\W$ has i.i.d.\@ entries.
We then treat the random feature matrix $\bPhi \equiv \phi(\W \X) \in \RR^{d \times n}$ and regress against the target $\y$. 
This leads to $\bbeta \in \RR^d$ minimizing the following ridge-regularized MSE
\begin{equation}\label{eq:RF_reg}
  L(\bbeta) =  \frac1{2n} \sum_{i=1}^n (y_i - \hat y(\x_i))^2 + \frac{\gamma}2 \| \bbeta\|_2^2  =  \frac1{2n} \| \y - \bPhi^\top \bbeta \|_2^2 + \frac{\gamma}2 \| \bbeta\|_2^2,
\end{equation}
where $\gamma \ge 0$ is the regularization penalty.
Its solution is uniquely given by
\begin{equation}\label{eq:def-RF_reg_beta}
  \bbeta_\gamma = \frac1n \bPhi \left( \frac1n \bPhi^\top \bPhi + \gamma \I_n \right)^{-1}\y = \left( \frac1n \bPhi \bPhi^\top + \gamma \I_d \right)^{-1} \frac1n \bPhi \y,
\end{equation}
for $\gamma > 0$. 
Hence, the training MSE is $E_{\train} = \frac1n \| \y - \bPhi^\top \bbeta_\gamma \|_2^2 = \frac{\gamma^2}n \frac{\partial \y^\top \Q(-\gamma) \y}{\partial \gamma}$, where
\begin{equation}\label{eq:def_nonlinear_Gram}
  \Q(- \gamma ) \equiv \left(\frac1n \bPhi^\top \bPhi + \gamma \I_n \right)^{-1}, \quad \bPhi^\top \bPhi = \phi(\X^\top \W^\top) \phi(\W \X) ,
\end{equation}
i.e., the \emph{resolvent} (\Cref{def:resolvent}) of the \emph{nonlinear Gram matrix} model $ \mathcal{M}_{\phi}(\X;\bTheta) = \frac1n \bPhi^\top \bPhi$.
As in the case of linear regression in \Cref{prop:risk_LS}, both $\Q$ and its Deterministic Equivalent $\tilde \Q$ (specified below in \Cref{theo:nonlinear-Gram}) are central to analyzing the \emph{nonlinear} NN model in \Cref{def:single-layer-NN}.

\begin{Remark}[\textbf{Nonlinear Gram matrix: classical versus proportional regime}]\label{rem:Nonlinear_Gram_classical_proportional}
\normalfont
Similar to the linear SCM discussed in \Cref{rem:SCM_classical_proportional}, the nonlinear Gram matrix $\bPhi^\top \bPhi$ in \eqref{eq:def_nonlinear_Gram} exhibits different behaviors.
\begin{enumerate}
  \item In the \emph{classical regime} for fixed $n,p$ and as $d \to \infty$, by the LLN, 
  \begin{align}
      \frac1d \bPhi^\top \bPhi = \frac1d \sum_{i=1}^d \phi(\X^\top \w_i) \phi(\w_i^\top \X) \to \EE_{\w}[\phi(\X^\top \w) \phi(\w^\top \X)] \equiv \K, \label{eq:def_K_NN}
  \end{align}
  almost surely, where $\w_i^\top \in \RR^{1 \times p}$ is the $i^{th}$ \emph{row} of $\W \in \RR^{d \times p}$, and the vectors $\w_i$ are i.i.d. 
  Thus, $\bPhi^\top \bPhi/d$ becomes nearly deterministic.
  \item In the \emph{proportional regime} as $n,p,d \to \infty$ at the same pace, $\bPhi^\top \bPhi/d$ does not converge to $\K$.
  Correspondingly, the resolvent $\Q(- \gamma )$ in \eqref{eq:def_nonlinear_Gram} is not deterministic, but it admits a Deterministic Equivalent (in the sense of \Cref{def:DE}) given in \Cref{theo:nonlinear-Gram} below.
\end{enumerate}
\end{Remark}

\ifisarxiv
\subsection{Deterministic Equivalent for resolvents of the nonlinear NN model}
\fi

We extend \Cref{theo:G_SCM_DE} (originally stated for linear Gram) to the nonlinear Gram matrix $\bPhi^\top \bPhi$. 

\begin{tcolorbox}[breakable]
\begin{Theorem}[\textbf{Deterministic Equivalent for nonlinear resolvent},~{\cite[Theorem~1]{louart2018random}}]\label{theo:nonlinear-Gram}
Let $\W \in \RR^{d \times p}$ be a random matrix with i.i.d.\@ sub-gaussian entries of zero mean and unit variance, and let $\X \in \RR^{p \times n}$ be independent of $\W$ with $\| \X \|_2 \leq 1$.
Define $\Q(z)$ as in \eqref{eq:def_nonlinear_Gram}, and let $\phi \colon \RR \to \RR$ be Lipschitz.
Then, for $z \in \CC \setminus (0,\infty)$ and as $n,p,d \to \infty$ together, the following Deterministic Equivalent holds
\begin{equation}\label{eq:def_bar_Q_NN}
  \Q(z) \overset{f}{\leftrightarrow} \tilde\Q(z), \quad \tilde\Q(z) = \left( \frac{d}n \frac{\K}{1+\delta(z)} - z \I_n \right)^{-1},
\end{equation}
for $f$ being trace and bilinear forms as in \Cref{rem:scalar_func}.
Here, $\delta(z)$ is the unique Stieltjes transform solution to $\delta(z) = \frac1n \tr \K \tilde\Q(z)$ and $\K$ is the \emph{kernel matrix} defined in \eqref{eq:def_K_NN} such that $\K \succ \zo$.
\end{Theorem}
\end{tcolorbox}

Analogous to \Cref{prop:risk_LS} for linear regression, \Cref{theo:nonlinear-Gram} can be used to derive the asymptotic training and test MSEs for the single-hidden-layer NN model in the proportional regime, as follows.

\begin{tcolorbox}[breakable]
\begin{Proposition}[\textbf{Asymptotic training and test MSEs for single-hidden-layer NNs in the proportional regime},~\cite{louart2018random}]\label{prop:NN-MSEs}
Consider the single-hidden-layer NN model in \Cref{def:single-layer-NN} under the setting and notation in \Cref{theo:nonlinear-Gram}, with $\tilde \Q(-\gamma) = \tilde \Q$ and $\delta(-\gamma) = \delta$.
Then, the training and test MSEs satisfy $E_{\train} - \tilde E_{\train} \to 0, E_{\test} - \tilde E_{\test} \to 0$, almost surely as $n,p,d \to \infty$, where $\tilde \K = \frac{d}n \frac{\K}{1 + \delta}$ and
\begin{align*}
    \tilde E_{\train} & = \frac{\gamma^2}n \y^\top \tilde \Q \left( \frac{ \tr (\tilde \Q \tilde \K \tilde \Q) }{ d - \tr (\tilde \K \tilde \Q \tilde \K \tilde \Q ) } \tilde \K + \I_n \right) \tilde \Q \y,\\ 
    \tilde E_{\test} & = \frac1{n'} \| \y' - \tilde \K^\top_{\X \X'} \tilde \Q \y \|_2^2 + \frac{ \y^\top \tilde \Q \tilde \K \tilde \Q \y }{ d - \tr (\tilde \K \tilde \Q \tilde \K \tilde \Q ) }  \frac1{n'} \left( \tr \tilde \K_{\X' \X'} - \tr \tilde \K_{\X \X'}^\top \tilde \Q (\I_n + \gamma \tilde \Q) \tilde \K_{\X \X'} \right).
\end{align*}
Here, $\K \equiv \K_{\X \X} = \EE_\w[\phi(\X^\top \w) \phi(\w^\top \X)]$ as in \eqref{eq:def_K_NN}, and similarly $\K_{\X \X'} = \EE_\w[\phi(\X^\top \w) \phi(\w^\top \X')] \in \RR^{n \times n'}, \K_{\X' \X'} = \EE_\w[\phi((\X')^\top \w) \phi(\w^\top \X')] \in \RR^{n' \times n'} $, and $\tilde \K_{\X \X'} \equiv \frac{d}n \frac{\K_{\X \X'}}{1 + \delta}, \tilde \K_{\X' \X'} \equiv \frac{d}n \frac{\K_{\X' \X'}}{1 + \delta}$.
\end{Proposition}
\end{tcolorbox}

\begin{figure}[htb]
\centering
\begin{subfigure}[t]{0.48\linewidth}
\centering
\begin{tikzpicture}[font=\footnotesize]
    \pgfplotsset{every major grid/.style={style=densely dashed}}
    \begin{axis}[
      height=.7\linewidth,
      width=.95\linewidth,
      xmin=1e-3,
      xmax=1,
      ymin=1e-4,
      ymax=1,
      xmode=log,
      ymode=log,
      grid=major,
      scaled ticks=true,
      xlabel={ $d/n$ },
      ylabel={ Training MSEs },
      legend style = {at={(0,0)}, anchor = south west, font=\footnotesize},
      ]
      \addplot[BLUE,mark=o,only marks,line width=1pt] coordinates{
      (0.000977,0.918195)(0.000977,0.920790)(0.000977,0.947885)(0.001953,0.874339)(0.001953,0.855032)(0.002930,0.799988)(0.003906,0.757640)(0.004883,0.696596)(0.005859,0.655987)(0.007812,0.578417)(0.009766,0.514627)(0.012695,0.459650)(0.016602,0.398307)(0.021484,0.327016)(0.027344,0.292008)(0.034180,0.251839)(0.043945,0.219120)(0.055664,0.189967)(0.070312,0.162421)(0.089844,0.135807)(0.114258,0.122572)(0.144531,0.098461)(0.183594,0.083106)(0.233398,0.066754)(0.295898,0.051989)(0.375977,0.039209)(0.477539,0.026666)(0.606445,0.015040)(0.769531,0.006199)(0.976562,0.000404)
      };
      \addlegendentry{{ $E_{\train}, \gamma = 10^{-5}$ }};
      \addplot[BLUE,densely dashed,line width=1pt] coordinates{
      (0.000977,0.919486)(0.000977,0.919486)(0.000977,0.919486)(0.001953,0.845757)(0.001953,0.845757)(0.002930,0.780482)(0.003906,0.723571)(0.004883,0.674175)(0.005859,0.631253)(0.007812,0.560967)(0.009766,0.506291)(0.012695,0.444192)(0.016602,0.385127)(0.021484,0.333768)(0.027344,0.290866)(0.034180,0.255455)(0.043945,0.220073)(0.055664,0.190624)(0.070312,0.164671)(0.089844,0.140346)(0.114258,0.118996)(0.144531,0.100208)(0.183594,0.082921)(0.233398,0.067148)(0.295898,0.052816)(0.375977,0.039359)(0.477539,0.026803)(0.606445,0.015353)(0.769531,0.006027)(0.976562,0.000575)
      };
      \addlegendentry{{ $\tilde E_{\train}, \gamma = 0$ }};
      \addplot[RED,mark=triangle,only marks,line width=1pt] coordinates{
      (0.000977,0.942436)(0.000977,0.950333)(0.000977,0.943024)(0.001953,0.845260)(0.001953,0.866893)(0.002930,0.781947)(0.003906,0.752737)(0.004883,0.684115)(0.005859,0.647850)(0.007812,0.593877)(0.009766,0.508103)(0.012695,0.434178)(0.016602,0.407565)(0.021484,0.311958)(0.027344,0.278762)(0.034180,0.240444)(0.043945,0.204988)(0.055664,0.175617)(0.070312,0.155783)(0.089844,0.133370)(0.114258,0.110268)(0.144531,0.097993)(0.183594,0.081723)(0.233398,0.068283)(0.295898,0.058782)(0.375977,0.048753)(0.477539,0.040249)(0.606445,0.032751)(0.769531,0.026642)(0.976562,0.021130)
      };
      \addlegendentry{ { $E_{\train}, \gamma = 10^{-1}$ } };
      \addplot[RED,densely dashed,line width=1pt] coordinates{
      (0.000977,0.915070)(0.000977,0.915070)(0.000977,0.915070)(0.001953,0.839121)(0.001953,0.839121)(0.002930,0.772601)(0.003906,0.714890)(0.004883,0.664930)(0.005859,0.621588)(0.007812,0.550738)(0.009766,0.495738)(0.012695,0.433419)(0.016602,0.374339)(0.021484,0.323179)(0.027344,0.280659)(0.034180,0.245774)(0.043945,0.211192)(0.055664,0.182707)(0.070312,0.157920)(0.089844,0.135046)(0.114258,0.115347)(0.144531,0.098395)(0.183594,0.083220)(0.233398,0.069868)(0.295898,0.058299)(0.375977,0.048075)(0.477539,0.039159)(0.606445,0.031412)(0.769531,0.024759)(0.976562,0.019110)
      };
      \addlegendentry{{ $\tilde E_{\train}, \gamma = 10^{-1}$ }};
    \end{axis}
  \end{tikzpicture}
  \caption{{Training MSE}} \label{subfig:NN_train}
\end{subfigure}
\hfill{}
\begin{subfigure}[t]{0.48\linewidth}
\centering
\begin{tikzpicture}[font=\footnotesize]
    \pgfplotsset{every major grid/.style={style=densely dashed}}
    \begin{axis}[
      height=.7\linewidth,
      width=.95\linewidth,
      xmin=0,
      xmax=2,
      ymin=0,
      ymax=2,
      grid=major,
      scaled ticks=true,
      xlabel={ $d/n$ },
      ylabel={ Test MSEs },
      legend style = {at={(1,1)}, anchor = north east, font=\footnotesize},
      ]
      \addplot[BLUE,mark=o,only marks,line width=1pt] coordinates{
          (0.000977,0.939868)(0.072266,0.174539)(0.143555,0.129443)(0.214844,0.119569)(0.286133,0.113666)(0.357422,0.117482)(0.428711,0.125761)(0.500000,0.138240)(0.571289,0.160027)(0.642578,0.180032)(0.713867,0.224658)(0.785156,0.269396)(0.856445,0.334971)(0.927734,0.524109)(1.000000,1.728754)(1.009766,0.683946)(1.080078,0.408252)(1.151367,0.275812)(1.221680,0.194984)(1.291992,0.163972)(1.363281,0.134866)(1.433594,0.123883)(1.504883,0.110242)(1.575195,0.105126)(1.645508,0.095413)(1.716797,0.092876)(1.787109,0.089558)(1.858398,0.083428)(1.928711,0.082909)(2.000000,0.079977)
      };
      \addlegendentry{{ $E_{\test}, \gamma = 10^{-5}$ }};
      \addplot[BLUE,densely dashed,line width=1pt] coordinates{
          (0.000977,0.915632)(0.072266,0.175597)(0.143555,0.133315)(0.214844,0.119946)(0.286133,0.116851)(0.357422,0.119823)(0.428711,0.127704)(0.500000,0.140477)(0.571289,0.158774)(0.642578,0.183845)(0.713867,0.218083)(0.785156,0.267159)(0.856445,0.347616)(0.927734,0.514055)(1.000000,2.731984)(1.009766,0.708555)(1.080078,0.403522)(1.151367,0.259712)(1.221680,0.196180)(1.291992,0.161140)(1.363281,0.138877)(1.433594,0.123885)(1.504883,0.112859)(1.575195,0.104618)(1.645508,0.098151)(1.716797,0.092876)(1.787109,0.088601)(1.858398,0.084977)(1.928711,0.081944)(2.000000,0.079302)
      };
      \addlegendentry{{ $\tilde E_{\test}, \gamma \to 0$ }};
      \addplot[RED,mark=triangle,only marks,line width=1pt] coordinates{
      (0.000977,0.935824)(0.072266,0.173081)(0.143555,0.128324)(0.214844,0.107916)(0.286133,0.095110)(0.357422,0.090767)(0.428711,0.085425)(0.500000,0.081491)(0.571289,0.077252)(0.642578,0.074521)(0.713867,0.072795)(0.785156,0.071466)(0.856445,0.070138)(0.927734,0.068992)(1.000000,0.068613)(1.009766,0.068396)(1.080078,0.066440)(1.151367,0.066102)(1.221680,0.065655)(1.291992,0.065524)(1.363281,0.065661)(1.433594,0.064201)(1.504883,0.063383)(1.575195,0.063277)(1.645508,0.063021)(1.716797,0.062769)(1.787109,0.062571)(1.858398,0.061712)(1.928711,0.061132)(2.000000,0.061558)
      };
      \addlegendentry{ { $E_{\test}, \gamma = 10^{-1}$ } };
      \addplot[RED,densely dashed,line width=1pt] coordinates{
      (0.000977,0.918837)(0.072266,0.179081)(0.143555,0.129704)(0.214844,0.109247)(0.286133,0.097709)(0.357422,0.090226)(0.428711,0.084959)(0.500000,0.081045)(0.571289,0.078018)(0.642578,0.075607)(0.713867,0.073641)(0.785156,0.072005)(0.856445,0.070623)(0.927734,0.069439)(1.000000,0.068400)(1.009766,0.068270)(1.080078,0.067401)(1.151367,0.066621)(1.221680,0.065935)(1.291992,0.065319)(1.363281,0.064755)(1.433594,0.064250)(1.504883,0.063783)(1.575195,0.063361)(1.645508,0.062972)(1.716797,0.062609)(1.787109,0.062276)(1.858398,0.061963)(1.928711,0.061675)(2.000000,0.061402)
      };
      \addlegendentry{{ $\tilde E_{\test}, \gamma = 10^{-1}$ }};
    \end{axis}
  \end{tikzpicture}
  \caption{{Test MSE}} \label{subfig:NN_test}
  \end{subfigure}
\caption{{
Empirical and theoretical training and test MSEs of single-hidden-layer NN model in \Cref{def:single-layer-NN} as a function of $d/n$, for $\gamma = 10^{-1}$ and $\gamma = 10^{-5}$, with Gaussian $\W$ and ReLU activation $\phi(t) = \max(t,0)$, $n = 1\,024$ training samples and $n' = 1\,024$ test samples from the MNIST digits 1 and 2.
\textbf{\Cref{subfig:NN_train}}: log-log plot of training MSEs averaged over $30$ runs.
\textbf{\Cref{subfig:NN_test}}: test MSEs averaged over $30$ runs on independent test sets.
}}
\label{fig:NN-reg}
\end{figure}

A few remarks regarding the \emph{nonlinear} single-hidden-layer NN model are in order.
\begin{Remark}[\textbf{Scaling law of training MSE}]\label{rem:scaling_train_NN}\normalfont
Similar to \Cref{rem:scaling_in_sample}, setting $\gamma = 0$ in \Cref{prop:NN-MSEs} and considering the under-parameterized regime with $n,p,d$ all large but $d < n$, we have that $\delta$ \emph{diverges} as $\gamma\to 0$. 
However, with $\rho \equiv d/n$, $\gamma \delta = \frac1n \tr \K \left( \rho \frac{\K}{\gamma + \gamma \delta} + \I_n \right)^{-1} \xrightarrow{\gamma \to 0} \theta = \frac1n \tr \K \left( \rho \frac{\K}{\theta} + \I_n \right)^{-1}$.
Equivalently, if $\lambda_1\geq\cdots\geq\lambda_n>0$ are the ordered eigenvalues of $\K$, then $\theta$ is characterized by $\rho=\frac1n\sum_{j=1}^n\frac{\lambda_j}{\lambda_j+\theta/\rho}$.
Further, if $\y =  \K^{1/2} \bbeta_*$ for $\bbeta_* \sim \NN(\zo,\I_n)$, then it follows from \Cref{prop:NN-MSEs} that $E_{\train} - \tilde E_{\train} \to 0$ with $\tilde E_{\train} \xrightarrow{\gamma \to 0} \theta$.
This implies \emph{explicit} scaling laws for the training MSEs that depend on the ordered eigenspectrum of $\K$.
For instance, polynomial eigendecay (e.g., Matérn kernel associated with ReLU activation~\cite{geifman2020similarity}), $\lambda_j=Cj^{-(1+\beta)}(1+o(1))$ with $\beta>0$, yields $\tilde E_{\train}\sim \rho b_{\rho}n^{-1-\beta}$ when $d/n\to\rho\in(0,1)$, where $b_{\rho}>0$ is determined by $\rho=\int_0^1(1+b_{\rho}x^{1+\beta})^{-1}dx$.
This gives a rate faster than the $n^{-1}$ rate of linear models in \Cref{rem:scaling_in_sample}.
\ifisarxiv
See \Cref{sec:proof_of_rem:scaling_train_NN} for a detailed proof.
\fi 
\end{Remark}

\begin{Remark}[\textbf{Double descent behavior for test MSE}]\label{rem:double_descent_test_NN}
\normalfont
Similar to the out-of-sample risk $R_{\rm out}$ for linear models discussed in \Cref{rem:double_descent_out-of-sample}, it follows from \Cref{prop:NN-MSEs} and the discussion in \Cref{rem:scaling_train_NN} that both $\theta$ and $\delta$ diverge as $\gamma \to 0$ at $n/d = 1$. 
Thus, the test risk likewise exhibits a singularity at $n/d = 1$.
This mirrors the double descent phenomenon for linear models (\Cref{rem:double_descent_out-of-sample}), but it applies here to nonlinear NN models, \emph{regardless of} the activation function \emph{or} the training/test data.\footnote{A counterexample for such divergence and singularity arises if $\X' = \X$ and $\y' = \y$.
In that case, the test error $\tilde E_{\test}$ reduces to the training error $\tilde E_{\train}$. 
Generally, however, when $\X'$ differs sufficiently from $\X$ (when measured by the associated  kernel matrices and when compared to $\gamma$), the test error diverges at $d=n$ as $\gamma\to 0$.
See \cite{liao2020random} for a detailed discussion.
}
\end{Remark}

\ifisarxiv
\subsection{High-dimensional linearization of single-hidden-layer NNs}
\fi

We now delve more deeply into how the nonlinear activation $\phi(\cdot)$ influences NN performance, by applying the high-dimensional linearization approach (based on Hermite polynomials in \Cref{theo:normalized_Hermite}).

From \Cref{theo:nonlinear-Gram} and \Cref{prop:NN-MSEs}, we see that the NN performance depends on the interaction of the data $\X$ and the activation $\phi(\cdot)$ \emph{only} through the nonlinear kernel matrix model $ \mathcal{M}_{\phi}(\X) =  \K = \EE_\w[\phi(\X^\top \w) \phi(\w^\top \X)]$ defined in \eqref{eq:def_K_NN}.
\ifisarxiv
The following result (proven in \Cref{sec:proof_of_theo:linearization_kernel}) shows how this kernel matrix can be linearized.
\fi

\begin{tcolorbox}[breakable]
\begin{Theorem}[\textbf{High-dimensional linearization of kernel matrix}]\label{theo:linearization_kernel}
Let $\w \in \RR^p$ be standard Gaussian $\w \sim \NN(\zo,\I_p)$ and let $\x_1, \ldots, \x_n \in \RR^p$ be independently and uniformly drawn from the unit sphere $\mathbb{S}^{p-1} \subset \RR^p$. 
Then, as $n,p \to \infty$ with $p/n \to c \in (0, \infty)$, the kernel matrix $\K = \EE_\w[\phi(\X^\top \w) \phi(\w^\top \X)]$ defined in \eqref{eq:def_K_NN} admits the following Linear Equivalent (see \Cref{def:Hi-LE}):
\begin{equation}
  \K \overset{f}{\leftrightarrow} \tilde \K_\phi, \quad \tilde \K_\phi = a_{\phi;0}^2 \one_n \one_n^\top + a_{\phi;1}^2 \X^\top \X + a_{\phi;2}^2 \cdot \frac1p \one_n \one_n^\top + \left(\nu_\phi - a_{\phi;0}^2 - a_{\phi;1}^2 \right) \I_n,
\end{equation}
for functions $f \colon \RR^{n \times n} \to \RR$ with a bounded Lipschitz constant with respect to matrix spectral norm, i.e., $|f(\A) - f(\B)| \leq C \| \A - \B \|_2,\forall \A, \B \in \RR^{n \times n}$ and some $C \in (0, \infty)$, where $a_{\phi;0}, a_{\phi;1}, a_{\phi;2}, \nu_{\phi}$ are the Hermite coefficients of $\phi$, as defined in \Cref{theo:normalized_Hermite}.
\end{Theorem}
\end{tcolorbox}

A striking (and perhaps counterintuitive) consequence of \Cref{theo:linearization_kernel} is that, in the proportional regime with $n,p$ both large and comparable, the eigenvalue distribution of $\K$ becomes \emph{independent} of the activation function $\phi$, up to a scaling and shift, for random and unstructured input data.
Specifically, when the input data are \emph{unstructured} and independently uniformly distributed on the unit sphere, the eigenspectrum of $\K$ coincides with that of $\X^\top \X$, which approximates the Mar{\u c}enko-Pastur law, and depends only on the dimension ratio $p/n$.
See~\cite{liao2021sparse} for a related application to sparse and quantized spectral clustering on structured and non-isotropic input data.

\begin{Remark}[\textbf{Learning dynamics of nonlinear NNs}]\label{rem:dynamics_NN}\normalfont
We should note that, beyond the static generalization performance discussed above, one may also use the introduced techniques to study the \emph{learning dynamics} of a nonlinear NN~\cite{liao2018dynamics,advani2020high,paquette2023Halting}.
Consider training the second-layer weights $\bbeta$ with gradient descent on the loss $L(\bbeta)$ in \eqref{eq:RF_reg}, i.e., $\bbeta(t) = \bbeta(t-1) - \eta \nabla_{\bbeta} L(\bbeta(t-1))$, where $\eta > 0$ denotes the \emph{learning rate}.\footnote{For clarity of presentation, we assume that $\bPhi$ has full row rank so that $\bPhi \bPhi^\top$ is invertible. Additionally, we omit the regularization term in $L(\bbeta)$ by setting $\gamma =0$.}
In continuous time (i.e., the gradient flow approximation), this procedure follows the ordinary differential~equation:
\begin{equation}\label{eq:gradient_flow_NN}
  \frac{d}{dt} \bbeta(t)  = - \eta \nabla_{\bbeta} L(\bbeta(t)) = - \left( \frac1n \bPhi \bPhi^\top \right) \bbeta(t) + \frac1n \bPhi \y,
\end{equation}
whose solution is given by
\begin{equation}\label{eq:beta_t_NN}
  \bbeta(t) = \exp \left( - \eta t \Q_{\bPhi \bPhi^\top/n}^{-1}(0) \right) \bbeta(t=0) + \left( \I_d - \exp \left( - \eta t \Q_{\bPhi \bPhi^\top/n}^{-1}(0) \right) \right) \frac1n \Q_{\bPhi \bPhi^\top/n}(0) \bPhi \y,
\end{equation}
starting from some initial $\bbeta(t=0)$. 
Here, $\Q_{\bPhi \bPhi^\top/n}(z) \equiv \left(\frac1n \bPhi \bPhi^\top - z \I_d \right)^{-1}$ is the \emph{resolvent} of $\frac1n \bPhi \bPhi^\top$.
As $t \to \infty$,  $\bbeta(t)$ converges to $\bbeta_{\gamma = 0}$ in \eqref{eq:def-RF_reg_beta}, i.e., the global minimizer of $L(\bbeta)$ for $\gamma = 0$.

For any given $t > 0$ and deterministic $\vv \in \RR^d$, \Cref{theo:spectral_func_contour} implies that
\begin{equation*}
  \vv^\top \bbeta(t) = -\frac1{ 2 \pi {\jmath}} \oint_\Gamma \left(\exp \left( - \eta t z \right) \cdot \vv^\top \Q_{\bPhi \bPhi^\top/n}(z) \bbeta(0) + z \left(1 - \exp \left( - \eta t z \right) \right) \frac1n \vv^\top \Q_{\bPhi \bPhi^\top/n}(z) \bPhi \y \right) dz,
\end{equation*}
where $\Gamma$ is a positively-oriented simple closed contour enclosing all the eigenvalues of $\frac1n \bPhi \bPhi^\top$.
If $\tilde{\Q}_{\bPhi \bPhi^\top/n}(z) \leftrightarrow \Q_{\bPhi \bPhi^\top/n}(z)$ is a Deterministic Equivalent for $\Q_{\bPhi \bPhi^\top/n}(z)$, then $\vv^\top \bbeta(t)$ (viewed as a scalar eigenspectral functional; see \Cref{def:spectral_functional}) can be evaluated using the proposed RMT framework described here.
\end{Remark}

\ifisarxiv
\section{Beyond Single-hidden-layer NN Models}\label{sec:DNN}
\fi

In this section, we extend the previous single-hidden-layer analysis to multilayer DNN models.
Untrained, randomly initialized DNNs can be characterized via their Conjugate Kernel (CK), which is the covariance function associated with the neural Gaussian process~\cite{lee2018deep}.
Much like the single-hidden-layer case in \Cref{theo:linearization_kernel}, we will see that the CK matrices of random DNNs can also be linearized.

When training a DNN via gradient descent (as discussed above in \Cref{rem:dynamics_NN} for shallow NNs), it is often more convenient to work with the Neural Tangent Kernel (NTK)~\cite{jacot2018neural}—closely related to the CK—which can likewise be linearized. 
The NTK formulation provides access to both learning dynamics and generalization.
We start by describing the setup.

\begin{tcolorbox}[breakable]
\begin{definition}[\textbf{Nonlinear DNN models}]\label{def:DNN}
An $L$-layer DNN has weight matrices $\W_1\in \RR^{d_1 \times p}, \cdots, \W_L  \in \RR^{d_L \times d_{L-1}}$ and a readout vector $\bbeta \in \RR^{d_L}$.
Given an input $\x \in \RR^p$, the network output is given by $\hat y(\x) = \bbeta^\top \phi_L\left( \W_L \phi_{L-1}\left( \ldots \phi_1 \left(\W_1 \x \right) \right)\right)$, 
where $\phi_1(\cdot), \ldots, \phi_L(\cdot)$ are entrywise activation functions applied at each layer.
For an input data matrix $\X = [\x_1, \ldots, \x_n] \in \RR^{p \times n}$, let $\bPhi_\ell(\X) \in \RR^{d_\ell \times n}$ be the intermediate representation at layer $\ell$:
\begin{equation}\label{eq:def_bSigma}
    \bPhi_\ell = \phi_\ell\left( \W_\ell \phi_{\ell-1}\left(\ldots \phi_2\left( \W_2 \phi_1\left(\W_1 \X\right)\right)\right)\right) \in \RR^{d_\ell \times n}, \quad \ell = 1, \ldots, L.
\end{equation}
For $\ell= 1$, $\bPhi_1$ coincides with the feature matrix $\bPhi$ of the single-hidden-layer NN in \Cref{def:single-layer-NN}.
\end{definition}
\end{tcolorbox}

Similar to the single-hidden-layer NN model, we first examine an untrained depth-$L$ DNN with random weights $\W_\ell \in \RR^{d_\ell \times d_{\ell - 1}}$ having i.i.d.\@ $\NN(0, 1/d_{\ell-1})$ entries, $\ell = 2, \ldots, L$ with standard Gaussian $\W_1$.
In the \emph{ultra-wide regime} with $d_\ell \gg \max(n,p)$, we have, by the LLN, that 
\begin{equation}\label{eq:def_K_ell}
  \frac1{d_\ell} \bPhi_\ell^\top \bPhi_\ell \to \frac1{d_\ell} \EE[\bPhi_\ell^\top \bPhi_\ell] \equiv \K_\ell,
\end{equation}
where the expectation is taken with respect to the random weights $\W_\ell, \W_{\ell - 1}, \ldots, \W_1$.
We refer to $\K_\ell$ in \eqref{eq:def_K_ell} as the \emph{Conjugate Kernel (CK)} matrix at layer $\ell$.
Notably, for $\ell = 1$, $\K_1$ recovers the kernel matrix $\K$ in \eqref{eq:def_K_NN} for single-hidden-layer NN.


We now extend the high-dimensional linearization result for single-hidden-layer NNs in \Cref{theo:linearization_kernel} to the CK of a DNN model.
\ifisarxiv
See \Cref{sec:proof_of_theo:CK} for the complete proof.
\fi

\begin{tcolorbox}[breakable]
\begin{Theorem}[\textbf{High-dimensional linearization of CK matrices for DNNs}]\label{theo:CK}
Consider a DNN as in \Cref{def:DNN}, with weights $\W_\ell \in \RR^{d_\ell \times d_{\ell - 1}}$ having i.i.d.\@ $\NN(0, 1/d_{\ell-1})$ entries for $\ell = 2, \ldots, L$ and standard Gaussian $\W_1$.
Assume each activation $\phi_\ell$ has Hermite coefficients (see \Cref{theo:normalized_Hermite}) satisfying $a_{\phi_\ell;0} = 0$ and $\nu_{\phi_\ell} = 1$.
Let $\x_1, \ldots, \x_n \in \RR^p$ be independently and uniformly drawn from the unit sphere $\mathbb{S}^{p-1} \subset \RR^p$. 
Then, as $n,p \to \infty$ with $p/n \to c \in (0, \infty)$, the CK matrix $\K_\ell = \EE[\bPhi_\ell^\top \bPhi_\ell]/d_\ell$ defined in \eqref{eq:def_K_ell} admits the following Linear Equivalent:
\begin{equation}
  \K_\ell \overset{f}{\leftrightarrow} \tilde \K_{\phi,\ell}, \quad \tilde \K_\phi = \alpha_{\ell,1}^2 \X^\top \X + \alpha_{\ell,2}^2 \cdot \frac1p \one_n \one_n^\top + \left( 1 - \alpha_{\ell,1}^2 \right) \I_n,
\end{equation}
for functions $f \colon \RR^{n \times n} \to \RR$ with a bounded Lipschitz constant with respect to matrix spectral norm, i.e., $|f(\A) - f(\B)| \leq C \| \A - \B \|_2,\forall \A, \B \in \RR^{n \times n}$ and some $C \in (0, \infty)$, for $\alpha_{\ell,1}, \alpha_{\ell,2} $ satisfying
\begin{align}
  \alpha_{\ell,1} &= a_{\phi_\ell;1} \cdot \alpha_{\ell-1,1}, \quad \alpha_{\ell,2} = \sqrt{ a_{\phi_\ell;1}^2 \cdot \alpha_{\ell-1,2}^2 + a_{\phi_\ell;2}^2 \cdot \alpha_{\ell-1,1}^4 }, \label{eq:def_ds2}
\end{align}
where $a_{\phi_\ell;1}, a_{\phi_\ell;2}$ are the Hermite coefficients of $\phi_\ell$ at layer $\ell$, as in \Cref{theo:normalized_Hermite}.
\end{Theorem}
\end{tcolorbox}

\begin{Remark}[\textbf{Deep versus shallow NNs and the ``curse of depth''}]\normalfont\label{rem:deep-versus-shallow}
Comparing \Cref{theo:CK} for DNNs to \Cref{theo:linearization_kernel} for single-hidden-layer NNs, we observe a ``curse of depth'' for random, untrained DNNs.
Specifically, since $\nu_{\phi_\ell} = \sum_{i = 0}^\infty a_{\phi_\ell;i}^2 = 1$ by \Cref{theo:normalized_Hermite}, we have $\max(a_{\phi_\ell;1}, a_{\phi_\ell;2}) \leq 1$ for each $\ell \in \{ 1, \ldots ,L \}$.
As a result, both $\alpha_{\ell,1}$ and $\alpha_{\ell,2}$ tend to decrease with growing depth $\ell$.

In particular, if $a_{\phi_\ell;1} < 1, \forall \ell \in \{1, \ldots, L\}$, then in the limit of $L \to \infty$, we obtain a \emph{degenerate} DNN with $\K_L \to \I_n$ that becomes independent of the input data.
This negative ``curse of depth'' result arises from:
\ifisarxiv
\begin{enumerate}
  \item \emph{unstructured} inputs ($\x$ uniformly distributed on the high-dimensional unit sphere); and 
  \item ``normalized'' activations ($a_{\phi_\ell;0} = 0$ and $\nu_{\phi_\ell} = 1$, $\forall \ell \in \{ 1, \ldots, L \}$); and
  \item random, untrained weights.
\end{enumerate}
\fi
In contrast with this, it has been shown in~\cite{gu2022Lossless} that for \emph{structured Gaussian mixture} inputs (which contain richer statistical information than the unstructured inputs considered in \Cref{theo:CK}), deeper (but only infinitely so, as shown in \cite{gu2022Lossless}) 
NNs with appropriately chosen activation functions can more effectively separate the input mixture, thereby outperforming their shallow counterparts.
See also~\cite{fan2020spectra} for a more practical, albeit weaker, result in the $n \sim p \sim d_\ell$ regime.
\end{Remark}

\begin{Remark}[\textbf{Learning dynamics of DNNs and the neural tangent kernel}]\label{rem:DNN_dynamics}\normalfont
We should note that we can extend the characterization of learning dynamics in \Cref{rem:dynamics_NN} for shallow NNs to the DNN model in \Cref{def:DNN}.
Rather than working with the (resolvent of the) Gram matrix $\frac1n \bPhi \bPhi^\top$ (or its linearized CK matrix $\K$ as in \Cref{theo:CK}), it is more convenient in the deep setting to work with the Neural Tangent Kernel (NTK)~\cite{jacot2018neural}, which naturally captures the gradient flow of the DNN output during training.

Denote $ \bTheta = [{\rm vec}(\W_1), \cdots, {\rm vec}(\W_L), \bbeta]$ the collection of all DNN weights. 
We consider the gradient flow that minimizes the square loss $L(\bTheta) =  \frac12 \| \y - \bPhi_L^\top \bbeta \|_2^2$ given by
\begin{equation}\label{eq:gradient_flow_DNN}
  \frac{d}{dt} \bTheta(t)  = - \eta \nabla_{\bTheta} L(\bTheta(t)),
\end{equation}
where $\nabla_{\bTheta} L(\bTheta(t)) $ is the gradient at time $t$, and $\eta$ is the learning rate.

The DNN output $\hat y(\X;t)$ at time $t$ then evolves according to
\begin{align}
    \frac{d}{dt} \hat y(\X;t)  = - \eta \cdot \nabla_{\bTheta} \hat y(\X;t) \cdot \nabla_{\bTheta} \hat y(\X;t )^\top \cdot \left(\hat y(\X;t) - \y \right) \equiv - \eta \cdot \K_{\NTK}(t) \cdot ( \hat y(\X;t) - \y), \label{eq:NTK_dynamic}
\end{align}
where we introduce the (symmetric p.s.d.\@) \emph{Neural Tangent Kernel} (NTK) matrix at time $t$ defined as
\begin{equation}\label{eq:def_K_NTK_t}
  \K_{\NTK}(t) \equiv \nabla_{\bTheta} \hat y(\X;t) \cdot \left(\nabla_{\bTheta} \hat y(\X;t) \right)^\top \in \RR^{n \times n}.
\end{equation}
In the \emph{ultra-wide regime} $d_\ell \gg \max(n,p)$, $\ell = 1, \cdots, L$, it has been shown in~\cite{jacot2018neural,lee2020wide} that the NTK matrix $\K_{\NTK}(t)$ in \eqref{eq:def_K_NTK_t} is nearly independent of $t$ and close to its initialization-time expectation, i.e.,
\begin{equation}\label{eq:NTK_approx}
  \K_{\NTK}(t > 0) \approx \K_{\NTK}(t=0) \approx \EE[\K_{\NTK}(t=0)] \equiv \K_{\NTK},
\end{equation}
in a Frobenius (and thus spectral) norm sense, up to an error of order $O(1/\sqrt{d_\ell})$.
In this regime, the solution to \eqref{eq:NTK_dynamic} can then be written explicitly as
\begin{equation}
    \hat y(\X;t) = \exp \left(- \eta t \cdot \K_{\NTK} \right) \hat y(\X;t=0) + \left( \I_n - \exp \left(- \eta  t \cdot \K_{\NTK} \right) \right) \y.
\end{equation}
This is analogous to the single-hidden-layer gradient-flow solution in \eqref{eq:beta_t_NN}, but now governed by $\K_{\NTK}$.

For a given (random) DNN, the deterministic NTK matrix $\K_{\NTK}$ in \eqref{eq:NTK_approx} is related to its CK matrix $\K_{\ell}$ in \eqref{eq:def_K_ell} via the recursion
\begin{equation}\label{eq:CK_NTK_recursion}
    \K_{\NTK, \ell}=\K_{\ell}+\K_{\NTK, \ell-1} \circ \K_{\ell}^{\prime}, \quad \ell = 1, \ldots, L,
\end{equation}
with $\K_{\NTK, 0}=\K_0 =\X^{\top} \X$, $\K_{\NTK} = \K_{\NTK,L}$, and `$\A \circ \B$' denoting the Hadamard product. 
Here, $\K'_{\ell}$ is defined similarly to $\K_{\ell}$ in \eqref{eq:def_K_ell} but using the derivative $\phi'_\ell$ in place of $\phi_\ell$ when forming $\bPhi_{\ell}$ in~\eqref{eq:def_bSigma}.

By applying the same linearization idea in \Cref{theo:CK} to the NTK matrix $\K_{\NTK}$ in the recursion \eqref{eq:CK_NTK_recursion}, the RMT-based characterization of gradient descent dynamics and generalization for single-hidden-layer NNs in \Cref{rem:dynamics_NN} can be extended to DNN models—at least in the ultra-wide NTK regime.
To the best of our knowledge, this perspective has not been fully explored in the literature. 
We believe it offers a promising future direction connecting RMT techniques with the learning dynamics of deep networks.
\end{Remark}

\ifisarxiv
\section{Conclusion}\label{sec:conclusion}
\fi

In this paper, we describe how to extend traditional RMT approaches beyond eigenvalue-based analysis of linear models to address nonlinear ML models of interest such as DNNs. 
We introduce the concept of High-dimensional Equivalent, which includes Deterministic Equivalent and Linear Equivalent as special cases. 
Using this framework, we derive precise characterizations of the training and generalization performance for linear, nonlinear shallow, and nonlinear deep neural networks. 
Our approach captures rich phenomena like scaling laws, double descent, and nonlinear learning dynamics---all behaviors that classical methods fail to address.
See again \Cref{fig:appetizer} for a high-level summary of these concepts and results.
Future work will focus on the integration of the RMT analysis methods we have presented with non-asymptotic RMT techniques, as combining these two approaches will provide a crucial tool for understanding and optimizing modern deep learning models.

\section*{Acknowledgment}

Z.~Liao would like to acknowledge the National Natural Science Foundation of China (via NSFC-62206101 and NSFC-12571561) and the Guangdong Provincial Key Laboratory of Mathematical Foundations for Artificial Intelligence (2023B1212010001) for providing partial support.
M.~W.~Mahoney would like to acknowledge the DARPA, DOE, NSF, and ONR for providing partial support of this work.

%
\bibliographystyle{IEEEtran}
\bibliography{liao}

@incollection{pastur2003Matrices,
  title = {{Matrices al{\'e}atoires: Statistique asymptotique des valeurs propres}},
  shorttitle = {{Matrices al{\'e}atoires}},
  booktitle = {{S{\'e}minaire de Probabilit{\'e}s XXXVI}},
  author = {Pastur, Leonid and Lejay, Antonie},
  editor = {Az{\'e}ma, Jacques and {\'E}mery, Michel and Ledoux, Michel and Yor, Marc},
  year = {2003},
  volume = {1801},
  pages = {135--164},
  publisher = {Springer Berlin Heidelberg},
  address = {Berlin, Heidelberg},
  doi = {10.1007/978-3-540-36107-7_2},
  urldate = {2025-06-06},
  isbn = {978-3-540-00072-3 978-3-540-36107-7},
}

@article{paquette2023Halting,
  title = {Halting {{Time}} Is {{Predictable}} for {{Large Models}}: {{A Universality Property}} and {{Average-Case Analysis}}},
  shorttitle = {Halting {{Time}} Is {{Predictable}} for {{Large Models}}},
  author = {Paquette, Courtney and {van Merri{\"e}nboer}, Bart and Paquette, Elliot and Pedregosa, Fabian},
  year = {2023},
  journal = {Foundations of Computational Mathematics},
  volume = {23},
  number = {2},
  pages = {597--673},
  issn = {1615-3383},
}

@inproceedings{mai2025the,
title={The Breakdown of Gaussian Universality in Classification of High-dimensional Linear Factor Mixtures},
author={Xiaoyi Mai and Zhenyu Liao},
booktitle={The Thirteenth International Conference on Learning Representations},
year={2025},
url={https://openreview.net/forum?id=UrKbn51HjA}
}

@book{rasmussenGaussianProcessesMachine2005,
  title = {Gaussian {{Processes}} for {{Machine Learning}}},
  author = {Rasmussen, Carl Edward and Williams, Christopher K. I.},
  year = {2005},
  month = nov,
  doi = {10.7551/mitpress/3206.001.0001}
}

@inproceedings{geifman2020similarity,
  title = {On the Similarity between the {Laplace} and {{Neural Tangent Kernels}}},
  booktitle = {Proceedings of the 34th {{International Conference}} on {{Neural Information Processing Systems}}},
  author = {Geifman, Amnon and Yadav, Abhay and Kasten, Yoni and Galun, Meirav and Jacobs, David and Basri, Ronen},
  year = {2020},
  pages = {1451--1461},
  publisher = {Curran Associates Inc.},
  urldate = {2025-02-13},
  isbn = {978-1-71382-954-6}
}

@incollection{vershynin2012introduction,
  title = {{Introduction to the Non-Asymptotic Analysis of Random Matrices}},
  booktitle = {Compressed Sensing: {{Theory}} and Applications},
  author = {Vershynin, Roman},
  editor = {Eldar, Yonina C. and Kutyniok, GittaEditors},
  year = {2012},
  pages = {210--268},
  publisher = {Cambridge University Press},
  doi = {10.1017/cbo9780511794308.006}
}

@inproceedings{montanari2022universality,
  title={Universality of empirical risk minimization},
  author={Montanari, Andrea and Saeed, Basil N},
  booktitle={Conference on Learning Theory},
  pages={4310--4312},
  year={2022},
  organization={PMLR}
}

@inproceedings{goldt2022gaussian,
  title={The gaussian equivalence of generative models for learning with shallow neural networks},
  author={Goldt, Sebastian and Loureiro, Bruno and Reeves, Galen and Krzakala, Florent and M{\'e}zard, Marc and Zdeborov{\'a}, Lenka},
  booktitle={Mathematical and Scientific Machine Learning},
  pages={426--471},
  year={2022},
  organization={PMLR}
}

@article{hu2022universality,
  title={Universality laws for high-dimensional learning with random features},
  author={Hu, Hong and Lu, Yue M.},
  journal={IEEE Transactions on Information Theory},
  volume={69},
  number={3},
  pages={1932--1964},
  year={2022},
  publisher={IEEE}
}

@article{couillet2016kernel,
  title = {Kernel Spectral Clustering of Large Dimensional Data},
  author = {Couillet, Romain and {Benaych-Georges}, Florent},
  year = {2016},
  journal = {Electronic Journal of Statistics},
  volume = {10},
  number = {1},
  pages = {1393--1454},
  issn = {1935-7524},
  doi = {10.1214/16-ejs1144}
}

@article{kammounCovarianceDiscriminativePower2023,
  title = {Covariance Discriminative Power of Kernel Clustering Methods},
  author = {Kammoun, Abla and Couillet, Romain},
  year = {2023},
  month = jan,
  journal = {Electronic Journal of Statistics},
  volume = {17},
  number = {1},
  pages = {291--390},
  publisher = {{Institute of Mathematical Statistics and Bernoulli Society}},
  issn = {1935-7524, 1935-7524},
  doi = {10.1214/23-EJS2107},
  urldate = {2023-04-05}
}

@article{tulino2004random,
  title = {Random Matrix Theory and Wireless Communications},
  author = {Tulino, Antonia M. and Verd{\'u}, Sergio},
  year = {2004},
  journal = {Foundations and Trends{\textregistered} in Communications and Information Theory},
  volume = {1},
  number = {1},
  pages = {1--182},
  issn = {1567-2190},
  doi = {10.1561/0100000001}
}

@inproceedings{gu2022Lossless,
  title = {{``Lossless'' Compression of Deep Neural Networks: A High-dimensional Neural Tangent Kernel Approach}},
  author = {Gu, Lingyu and Du, Yongqi and Zhang, Yuan and Xie, Di and Pu, Shiliang and Qiu, Robert and Liao, Zhenyu},
  volume = {35},
  pages = {3774--3787},
  booktitle = {Advances in Neural Information Processing Systems},
  publisher = {Curran Associates, Inc.},
  year = {2022},
}

@inproceedings{liao2021sparse, 
year = {2021}, 
author = {Liao, Zhenyu and Couillet, Romain and Mahoney, Michael W.}, 
title = {{Sparse Quantized Spectral Clustering}}, 
booktitle = {International Conference on Learning Representations}, 
url = {https://openreview.net/forum?id=pBqLS-7KYAF}
}

@article{rudelson2013hansonwright,
  title = {Hanson-{{Wright}} Inequality and Sub-Gaussian Concentration},
  author = {Rudelson, Mark and Vershynin, Roman},
  year = 2013,
  journal = {Electronic Communications in Probability},
  volume = {18},
  pages = {1--9},
  publisher = {{Institute of Mathematical Statistics and Bernoulli Society}},
  doi = {10.1214/ECP.v18-2865}
}

@inproceedings{lee2018deep, 
year = {2018}, 
author = {Lee, Jaehoon and Sohl-dickstein, Jascha and Pennington, Jeffrey and Novak, Roman and Schoenholz, Sam and Bahri, Yasaman}, 
title = {{Deep Neural Networks as Gaussian Processes}}, 
booktitle = {International Conference on Learning Representations}
}

@book{couillet2022RMT4ML,
  title={Random Matrix Methods for Machine Learning},
  author={Couillet, Romain and Liao, Zhenyu},
  year={2022},
  isbn={9781009186742},
  publisher={Cambridge University Press}
}

@inproceedings{liao2020random,
 author = {Liao, Zhenyu and Couillet, Romain and Mahoney, Michael W},
 booktitle = {Advances in Neural Information Processing Systems},
 pages = {13939--13950},
 publisher = {Curran Associates, Inc.},
 title = {A random matrix analysis of random Fourier features: beyond the Gaussian kernel, a precise phase transition, and the corresponding double descent},
 volume = {33},
 year = {2020}
}

@article{mei2021generalization, 
year = {2021}, 
title = {{The Generalization Error of Random Features Regression: Precise Asymptotics and the Double Descent Curve}}, 
author = {Mei, Song and Montanari, Andrea}, 
journal = {Communications on Pure and Applied Mathematics}, 
issn = {0010-3640}, 
doi = {10.1002/cpa.22008}
}

@article{bartlett2020benign, 
year = {2020}, 
title = {{Benign overfitting in linear regression}}, 
author = {Bartlett, Peter L. and Long, Philip M. and Lugosi, Gábor and Tsigler, Alexander}, 
journal = {Proceedings of the National Academy of Sciences}, 
issn = {0027-8424}, 
doi = {10.1073/pnas.1907378117}, 
pmid = {32332161}, 
pages = {30063--30070}, 
number = {48}, 
volume = {117}
}

@book{potters2020first, 
year = {2020}, 
title = {{A First Course in Random Matrix Theory: for Physicists, Engineers and Data Scientists}}, 
author = {Potters, Marc and Bouchaud, Jean-Philippe}, 
publisher = {Cambridge University Press}, 
doi = {10.1017/9781108768900}
}

@article{louart2018random, 
year = {2018}, 
title = {{A random matrix approach to neural networks}}, 
author = {Louart, Cosme and Liao, Zhenyu and Couillet, Romain}, 
journal = {Annals of Applied Probability}, 
issn = {1050-5164}, 
pages = {1190--1248}, 
number = {2}, 
volume = {28}
}

@article{bun2017cleaning, 
year = {2017}, 
title = {{Cleaning large correlation matrices: Tools from Random Matrix Theory}}, 
author = {Bun, Joël and Bouchaud, Jean-Philippe and Potters, Marc}, 
journal = {Physics Reports}, 
issn = {0370-1573}, 
doi = {10.1016/j.physrep.2016.10.005}, 
eprint = {1610.08104}, 
pages = {1--109}, 
volume = {666}
}

@book{pastur2011eigenvalue, 
year = {2011}, 
title = {{Eigenvalue Distribution of Large Random Matrices}}, 
author = {Pastur, Leonid Andreevich and Shcherbina, Mariya}, 
volume = {171}, 
series = {Mathematical Surveys and Monographs}, 
publisher = {American Mathematical Society}, 
doi = {10.1090/surv/1}
}

@article{dobriban2018high, 
year = {2018}, 
title = {{High-dimensional asymptotics of prediction: Ridge regression and classification}}, 
author = {Dobriban, Edgar and Wager, Stefan}, 
journal = {The Annals of Statistics}, 
issn = {0090-5364}, 
doi = {10.1214/17-aos1549}, 
pages = {247--279}, 
number = {1}, 
volume = {46}
}

@inproceedings{jacot2018neural, 
year = {2018}, 
author = {Jacot, Arthur and Gabriel, Franck and Hongler, Clément}, 
title = {{Neural Tangent Kernel: Convergence and Generalization in Neural Networks}}, 
booktitle = {Advances in Neural Information Processing Systems},
pages = {8571--8580}, 
volume = {31}, 
publisher = {Curran Associates, Inc.}
}

@inproceedings{liao2018dynamics, 
year = {2018}, 
author = {Liao, Zhenyu and Couillet, Romain}, 
title = {{The Dynamics of Learning: A Random Matrix Approach}}, 
booktitle = {Proceedings of the 35th International Conference on Machine Learning}, 
pages = {3072--3081}, 
volume = {80}, 
series = {Proceedings of Machine Learning Research}, 
publisher = {PMLR}, 
}

@article{bai2004clt, 
year = {2004}, 
title = {{CLT for linear spectral statistics of large-dimensional sample covariance matrices}}, 
author = {Bai, Zhidong and Silverstein, Jack W.}, 
journal = {The Annals of Probability}, 
issn = {0091-1798}, 
doi = {10.1214/aop/1078415845}, 
pages = {553--605}, 
number = {1A}, 
volume = {32}
}

@article{hastie2022Surprises,
  title = {Surprises in High-Dimensional Ridgeless Least Squares Interpolation},
  author = {Hastie, Trevor and Montanari, Andrea and Rosset, Saharon and Tibshirani, Ryan J.},
  year = {2022},
  month = apr,
  journal = {The Annals of Statistics},
  volume = {50},
  number = {2},
  pages = {949--986},
  publisher = {{Institute of Mathematical Statistics}},
  issn = {0090-5364, 2168-8966},
  doi = {10.1214/21-AOS2133},
  urldate = {2023-11-23}
}

@book{anderson2010introduction, 
year = {2010}, 
title = {{An Introduction to Random Matrices}}, 
author = {Anderson, Greg W. and Guionnet, Alice and Zeitouni, Ofer}, 
isbn = {9780511801334}, 
volume = {118}, 
series = {Cambridge Studies in Advanced Mathematics}, 
publisher = {Cambridge University Press}, 
doi = {10.1017/cbo9780511801334}
}

@book{andrews1999special, 
year = {1999}, 
title = {{Special Functions}}, 
author = {Andrews, George E. and Askey, Richard and Roy, Ranjan}, 
isbn = {9781107325937}, 
volume = {71}, 
series = {Encyclopedia of Mathematics and its Applications}, 
publisher = {Cambridge University Press}, 
address = {Cambridge}, 
doi = {10.1017/cbo9781107325937}
}

@article{belkin2019reconciling, 
year = {2019}, 
title = {{Reconciling modern machine-learning practice and the classical bias–variance trade-off}}, 
author = {Belkin, Mikhail and Hsu, Daniel and Ma, Siyuan and Mandal, Soumik}, 
journal = {Proceedings of the National Academy of Sciences}, 
issn = {0027-8424}, 
doi = {10.1073/pnas.1903070116}, 
pmid = {31341078}, 
pages = {15849--15854}, 
number = {32}, 
volume = {116}
}

@article{advani2020high, 
year = {2020}, 
title = {{High-dimensional dynamics of generalization error in neural networks}}, 
author = {Advani, Madhu S. and Saxe, Andrew M. and Sompolinsky, Haim}, 
journal = {Neural Networks}, 
issn = {0893-6080}, 
doi = {10.1016/j.neunet.2020.08.022}, 
pmid = {33022471}, 
pages = {428--446}, 
volume = {132}
}

@book{bai2010spectral, 
year = {2010}, 
title = {{Spectral Analysis of Large Dimensional Random Matrices}}, 
author = {Bai, Zhidong and Silverstein, Jack W.}, 
isbn = {9781441906601}, 
volume = {20}, 
series = {Springer Series in Statistics}, 
publisher = {Springer-Verlag New York}, 
edition = {2}, 
doi = {10.1007/978-1-4419-0661-8}
}

@article{marvcenko1967distribution, 
year = {1967}, 
title = {{Distribution of eigenvalues for some sets of random matrices}}, 
author = {Marcenko, Vladimir A and Pastur, Leonid Andreevich}, 
journal = {Mathematics of the USSR-Sbornik}, 
issn = {0025-5734}, 
doi = {10.1070/sm1967v001n04abeh001994}, 
pages = {457}, 
number = {4}, 
volume = {1}
}

@article{hachem2007deterministic, 
year = {2007}, 
title = {{Deterministic equivalents for certain functionals of large random matrices}}, 
author = {Hachem, Walid and Loubaton, Philippe and Najim, Jamal}, 
journal = {The Annals of Applied Probability}, 
issn = {1050-5164}, 
doi = {10.1214/105051606000000925}, 
eprint = {math/0507172}, 
pages = {875--930}, 
number = {3}, 
volume = {17}
}

@article{lee2020wide, 
year = {2020}, 
title = {{Wide neural networks of any depth evolve as linear models under gradient descent}}, 
author = {Lee, Jaehoon and Xiao, Lechao and Schoenholz, Samuel S and Bahri, Yasaman and Novak, Roman and Sohl-Dickstein, Jascha and Pennington, Jeffrey}, 
journal = {Journal of Statistical Mechanics: Theory and Experiment}, 
doi = {10.1088/1742-5468/abc62b}, 
pages = {124002}, 
number = {12}, 
volume = {2020}
}

@article{lytova2009central, 
year = {2009}, 
title = {{Central limit theorem for linear eigenvalue statistics of random matrices with independent entries}}, 
author = {Lytova, Anna and Pastur, Leonid}, 
journal = {The Annals of Probability}, 
issn = {0091-1798}, 
doi = {10.1214/09-aop452}, 
eprint = {0809.4698}, 
pages = {1778--1840}, 
number = {5}, 
volume = {37}
}

@article{tao2010random, 
year = {2010}, 
title = {{Random matrices: Universality of ESDs and the circular law}}, 
author = {Tao, Terence and Vu, Van and Krishnapur, Manjunath}, 
journal = {The Annals of Probability}, 
issn = {0091-1798}, 
doi = {10.1214/10-aop534}, 
pages = {2023--2065}, 
number = {5}, 
volume = {38}
}

@book{couillet2011random, 
year = {2011}, 
title = {{Random Matrix Methods for Wireless Communications}}, 
author = {Couillet, Romain and Debbah, Mérouane}, 
isbn = {9780511994746}, 
publisher = {Cambridge University Press}, 
doi = {10.1017/cbo9780511994746}
}

@inproceedings{fan2020spectra,
  title = {Spectra of the {{Conjugate Kernel}} and {{Neural Tangent Kernel}} for Linear-Width Neural Networks},
  booktitle = {Advances in Neural Information Processing Systems},
  author = {Fan, Zhou and Wang, Zhichao},
  year = {2020},
  volume = {33},
  pages = {7710--7721},
  publisher = {Curran Associates, Inc.}
}

\vfill

\ifisarxiv
\appendix


\section{Derivation of \Cref{theo:G_SCM_DE}}
\label{sec:proof_G_SCM_DE}

We present here the following (heuristic, and thus more accessible) derivation of \Cref{theo:G_SCM_DE}.
To simplify the derivation, we outline the proof of \Cref{theo:G_SCM_DE} by dropping the probabilistic controls on how scalar functionals $f(\Q)$ and quadratic forms of the type $\frac1n \x_i^\top \Q_{-i} \x_i$ in \eqref{eq:SM_quadratic_concentration} concentrate around their expectations. 
Such controls can be obtained using concentration inequalities such as Hanson-Wright~\cite{rudelson2013hansonwright}.
A detailed proof of \Cref{theo:G_SCM_DE} can be found in \cite[Section~2.2.2]{couillet2022RMT4ML}.

We first present in the following remark a general recipe for deriving Deterministic Equivalents.
\begin{Remark}[Deriving Deterministic Equivalents]\normalfont
\label{rem:proof_DE}
Mathematically, the derivation of a Deterministic Equivalent is generally accomplished via the following two steps:
\begin{enumerate}
  \item
  \textbf{Computing or approximating the expectation of the random matrix $\Q$.} 
  For the scalar functionals of interest $f(\Q)$ for $\Q \in \RR^{n \times n}$, the first (and often most natural) \emph{deterministic} quantity to describe its behavior is the expectation $\EE[f(\Q)]$.
  In the case of a linear or bilinear functional $f(\cdot)$ as in \Cref{def:DE} and \Cref{rem:scalar_func}, this is equal to $f(\EE[\Q])$. 
  In the case where $\EE[\Q]$ is not easily accessible, one may resort to approximating it using some deterministic matrix $\tilde \Q$, rather than directly computing it, e.g., find $\tilde \Q$ such that $\| \EE[\Q] - \tilde \Q \|_2 \to 0$ as $n \to \infty$.
  \item
  \textbf{Establishing the LLN-type concentration of $f(\Q)$ around $f(\tilde \Q)$.} 
  This step often involves concentration inequalities of the form
  \begin{equation}
    \Pr ( |f(\Q) - f(\tilde \Q) | \geq t) \leq \delta(n,t),
  \end{equation}
  for some function $\delta(n,t)$ that decreases to zero as $n \to \infty$. 
  This can be achieved, e.g., by bounding sequentially the differences $f(\Q) - f(\EE[\Q])$ and $f(\EE[\Q]) - f(\tilde \Q)$. 
  (The latter uses that the two \emph{deterministic} matrices $\EE[\Q]$ and $\tilde \Q$ are close in spectral norm, as established in the first step.)
\end{enumerate} 
\end{Remark}

We now derive the expressions of the Deterministic Equivalent $\tilde \Q_{\hat \C}$ in \eqref{eq:def_bar_Q_SCM} of \Cref{theo:G_SCM_DE}, following the recipe in \Cref{rem:proof_DE}.
We will show in the following that the difference in spectral norm $\| \EE[\Q_{\hat \C}] - \tilde \Q_{\hat \C} \|_2$ is small for some \emph{deterministic} matrix $\tilde \Q_{\hat \C}$ for $n,p$ large. 
We propose $\tilde \Q_{\hat \C} = \tilde \Q = (\F - z \I_p)^{-1}$ for some deterministic $\F \in \RR^{p \times p}$ to be determined, and we try to ``guess'' the form of such $\F$.
First, for $\Q_{\hat \C} = \Q$ and with the resolvent identity, we can write that 
\begin{align*}
    \EE[\Q - \tilde \Q] &= \EE\left[ \Q \left(\F - \frac1n \X \X^\top\right) \right] \tilde \Q \\ 
    &= \EE[\Q] \F \tilde \Q - \frac1n \sum_{i=1}^n \EE\left[ \Q \x_i \x_i^\top \right] \tilde \Q \\ 
    &= \EE[\Q] \F \tilde \Q - \frac1n \sum_{i=1}^n \EE\left[ \frac{\Q_{-i} \x_i \x_i^\top}{ 1+ \frac1n \x_i^\top \Q_{-i} \x_i } \right] \tilde \Q ,
\end{align*}
where we denote $\x_i = \C^{\frac12} \z_i \in \RR^p$ the $i^{th}$ column of $\X \in \RR^{p \times n}$, for $\z_i \in \RR^p$ the $i^{th}$ column of $\Z$, as well as $\Q_{-i} = (\frac1n \sum_{j \neq i} \x_j \x_j^\top - z \I_p)^{-1} = (\frac1n \X \X^\top - \frac1n \x_i \x_i^\top - z \I_p)^{-1}$ that is \emph{independent} of $\x_i$, and used the Sherman--Morrison formula in the third line.

Further note in the denominator that
\begin{equation}\label{eq:SM_quadratic_concentration}
  \frac1n \x_i^\top \Q_{-i} \x_i = \frac1n \z_i^\top \C^{\frac12} \Q_{-i} \C^{\frac12} \z_i \simeq \frac1n \tr \tilde \Q \C \equiv \delta,
\end{equation}
where we use $\simeq $ to approximate the quadratic form $\frac1n \z_i^\top \C^{\frac12} \Q_{-i} \C^{\frac12} \z_i$ by its expectation (using, e.g., Hanson-Wright inequality) 
and use the Deterministic Equivalent relations $\Q_{-i} \leftrightarrow \Q \leftrightarrow \tilde \Q$.

As such, we have
\begin{align*}
    \EE[\Q - \tilde \Q] &\simeq  \EE[\Q] \F \tilde \Q - \frac1n \sum_{i=1}^n \frac{ \EE\left[ \Q_{-i} \x_i \x_i^\top \right]}{ 1 + \delta }  \tilde \Q \\
    &= \EE[\Q] \F \tilde \Q - \frac1n \sum_{i=1}^n \frac{ \EE [ \Q_{-i}] \C }{ 1 + \delta }  \tilde \Q \\ 
    &\simeq  \EE[\Q] \left(\F - \frac{ \C }{ 1+ \delta }  \right) \tilde \Q,
\end{align*}
by exploiting the independence between $\Q_{-i}$ and $\x_i$ in the second line, and using the fact that $\| \EE[\Q - \Q_{-i}] \|_2 = O(n^{-1})$ in the third line. 
To have $\EE[\Q] \simeq \tilde \Q$, it thus suffices to take $\F$ such that the middle term vanishes. 

It then remains to derive self-consistent equations for $\delta(z)$ as
\begin{align*}
  \delta(z) &= \frac1n \tr \left( \tilde \Q(z) \C \right).
\end{align*}

This leads to the expression of $\tilde \Q(z) = \tilde \Q_{\hat \C}(z)$ in \eqref{eq:def_bar_Q_SCM} of \Cref{theo:G_SCM_DE}. 

\medskip

To obtain the expression of $\tilde \Q_{\G}(z) $ in \eqref{eq:def_bar_Q_G}, note that
\begin{equation}
  \Q_{\G} \cdot \frac1n \X^\top \X = \Q_{\G} (\frac1n \X^\top \X - z \I_n + z \I_n) = \I_n + z \Q_{\G},
\end{equation}
so that $\Q_{\G} = (\frac1n \X^\top \X - z \I_n)^{-1}$ is connected to $\Q_{\hat \C} = (\frac1n \X\X^\top - z \I_p)^{-1} = \Q$ through
\begin{equation}
  \EE[\Q_{\G}] = \frac1z \frac1n \EE[\Q_{\G} \X^\top \X] - \frac1z \I_n = \frac1z \frac1n \EE[\X^\top \Q  \X] - \frac1z \I_n,
\end{equation}
where we used $\A (\B \A - z \I_n)^{-1} = (\A \B - z \I_p)^{-1} \A$ for $\A \in \RR^{p \times n}$ and $\B \in \RR^{n \times p}$, for $z \in \CC$ distinct from $0$ and from the eigenvalues of $\A \B$.

It thus suffices to evaluate the $(i,j)^{th}$ entry of the expressions on both sides. 
Note, for $\x_i = \C^{\frac12} \z_i$ that
\begin{equation*}
  \frac1n \EE [\x_i^\top \Q \x_i] = \EE \left[ \frac{ \frac1n \x_i^\top \Q_{-i} \x_i}{ 1 + \frac1n \x_i^\top \Q_{-i} \x_i } \right] \simeq 1 - \frac1{ 1 + \delta(z) },
\end{equation*}
and for $\x_j = \C^{\frac12} \z_j$, $j \neq i$, that
\begin{equation*}
  \frac1n \EE [\x_i^\top \Q \x_j] = \EE \left[ \frac{ \frac1n \x_i^\top \Q_{-ij} \x_j}{ (1 + \frac1n \x_i^\top \Q_{-i} \x_i ) (1 + \frac1n \x_j^\top \Q_{-ij} \x_j ) } \right] \simeq 0,
\end{equation*}
with $\Q_{-ij} = (\frac1n \X \X^\top - \frac1n \x_i \x_i^\top - \frac1n \x_j \x_j^\top - z \I_p )^{-1}$ that is \emph{independent} of both $\x_i$ and $\x_j$, by using the Sherman--Morrison formula twice, as well as the independence between $\x_i$ and $\x_j$. 
Since $\Q_{-i} \leftrightarrow \Q \leftrightarrow \tilde \Q$, putting these together in matrix form gives
\begin{equation}
  \EE[\Q_{\G}] \simeq -\frac1{z (1 + \delta(z))} \I_n.
\end{equation}
This leads to the expression of $\tilde \Q_{\G}(z) $ in \eqref{eq:def_bar_Q_G} of \Cref{theo:G_SCM_DE}. 

\section{Proof of \Cref{prop:risk_LS}}
\label{sec:proof_of_risk_LS}

Denote $\Q(-\gamma) \equiv (\hat \C + \gamma \I_p)^{-1}$ the resolvent of the SCM $\hat \C = \frac1n \X \X^\top$ as in \Cref{def:resolvent} and $\Q(\gamma=0) = \lim_{\gamma \downarrow 0} \Q(-\gamma)$.
It follows from \Cref{eq:def-linear_reg_beta} and \Cref{def:noisy_linear_model} that 
\begin{equation}
  \bbeta_\gamma = \left(\hat \C + \gamma \I_p \right)^{-1} \frac1n \X \y = \Q(-\gamma) \frac1n \X \y = \Q(-\gamma) \hat \C \bbeta_* + \Q(-\gamma) \frac1n \X \boldsymbol{\epsilon},
\end{equation}
for $\boldsymbol{\epsilon} = [\epsilon_1, \ldots, \epsilon_n] \in \RR^n$, so that the in-sample risk $R_{\rm in}(\bbeta_\gamma)$ and the out-of-sample risk $R_{\rm out}(\bbeta_\gamma)$ in \Cref{def:noisy_linear_model} satisfy
\begin{align*}
  R_{\rm in}(\bbeta_\gamma) &= \frac1n \EE[ \| \X^\top \bbeta_\gamma - \X^\top \bbeta_* \|_2^2~|~\X] = \frac1n \left\| \X^\top ( \I_p - \Q(-\gamma) \hat \C ) \bbeta_* \right\|_2^2 + \frac{\sigma^2}n \tr \left( \Q(-\gamma) \hat \C \Q(-\gamma) \hat \C \right)  \\ 
  R_{\rm out}(\bbeta_\gamma) &= \EE[(\bbeta_\gamma^\top \x' - \bbeta_*^\top \x')^2~|~\X] = \left\| ( \I_p - \Q(-\gamma) \hat \C ) \bbeta_* \right\|_2^2 + \frac{\sigma^2}n \tr \left( \Q(-\gamma) \hat \C \Q(-\gamma) \right). 
\end{align*}
For $\gamma > 0$, we have $\I_p - \Q(-\gamma) \hat \C = \I_p - \Q(-\gamma) (\hat \C + \gamma \I_p - \gamma \I_p) = \gamma \Q(-\gamma)$, so that
\begin{align}
  R_{\rm in}(\bbeta_\gamma) & = \gamma^2 \left(\bbeta_*^\top \Q(-\gamma) \bbeta_* + \gamma \frac{\partial \bbeta_*^\top \Q(-\gamma) \bbeta_*}{\partial \gamma} \right) + \sigma^2 \left( \frac{p}n - \frac{2 \gamma}n \tr \Q(-\gamma) - \frac{\gamma^2}n \frac{\partial \tr \Q(-\gamma)}{\partial \gamma} \right), \label{eq:R_in} \\  
  R_{\rm out}(\bbeta_\gamma) &= -\gamma^2 \frac{\partial \bbeta_*^\top \Q(-\gamma) \bbeta_*}{\partial \gamma} + \sigma^2 \left( \frac1n \tr \Q(-\gamma) + \frac{\gamma}n \frac{\partial \tr \Q(-\gamma)}{\partial \gamma} \right), \label{eq:R_out}
\end{align}
where we used the fact that $\partial \Q(-\gamma)/\partial \gamma = - \Q^2(- \gamma)$. 
It thus suffices to evaluate the quadratic and trace forms of the random resolvent matrix $\Q(-\gamma)$.

\subsection{Characterization in the classical limit}

In the classical regime with $n \gg p$, we appeal to the law of large numbers to show $\hat \C \to \I_p$ almost surely for fixed $p$ and as $n \to \infty$, and therefore
\begin{equation}
  \Q(-\gamma) \to (\C + \gamma \I_p)^{-1} = \frac{\I_p}{1 + \gamma}.
\end{equation}
Thus, it follows from \Cref{eq:R_in} and \Cref{eq:R_out} that
\begin{align*}
  R_{\rm in}(\bbeta_\gamma) & - R_{{\rm in}, n \gg p}(\gamma) \to 0, \\ 
  R_{\rm out}(\bbeta_\gamma) & - R_{{\rm out}, n \gg p}(\gamma) \to 0. 
\end{align*}
with
\begin{equation}
  R_{{\rm in}, n \gg p}(\gamma) = R_{{\rm out}, n \gg p}(\gamma) = \frac{\gamma^2}{(1+\gamma)^2} \| \bbeta_* \|_2^2 + \frac{p}n \frac{\sigma^2}{ (1+\gamma)^2 }.
\end{equation}

\paragraph{Ridgeless case.}
In the ridgeless setting with $\gamma = 0$, we have $\I_p - \Q(\gamma=0) \hat \C = \I_p - \hat \C^+ \hat \C$, which is the projection onto $\ker(\hat \C)$. 
If $\hat \C$ is invertible (which is almost surely the case since $\hat \C \to \I_p$), then $\I_p - \Q(\gamma=0) \hat \C = \zo$ and $\Q(\gamma = 0) = \I_p$, so that 
\begin{equation}
  R_{{\rm in}, n \gg p}(\bbeta_0) = R_{{\rm out}, n \gg p}(\bbeta_0) = \sigma^2 \frac{p}n.
\end{equation}
This concludes the proof of \Cref{prop:risk_LS} in the classical $n \gg p$ limit, for any $\gamma \geq 0$.

\subsection{Characterization in the proportional limit}
We next consider the proportional regime with $n,p$ both large.
In this case, the SCM $\hat \C = \frac1n \X \X^\top$ is \emph{not} close to its population counterpart, and we do not expect the in-sample~and~out-of-sample risks in \Cref{def:noisy_linear_model} to yield the same behavior as in the classical regime, as in the first item of \Cref{prop:risk_LS}.
Instead, we prove this second item of \Cref{prop:risk_LS} by applying our \Cref{theo:G_SCM_DE} (and more precisely the special case of $\C = \I_p$ in \Cref{rem:MP}).

For $\gamma > 0$ and as $n,p \to \infty$ with $p/n \to c \in (0,\infty)$, it follows from \Cref{theo:G_SCM_DE}, \Cref{eq:R_in} and \Cref{eq:R_out} that
\begin{align*}
  R_{\rm in}(\bbeta_\gamma) & - R_{{\rm in}, n \sim p}(\gamma) \to 0, \\ 
  R_{\rm out}(\bbeta_\gamma) &- R_{{\rm out}, n \sim p}(\gamma) \to 0,
\end{align*}
with
\begin{align*}
  R_{{\rm in}, n \sim p}(\gamma) &= \gamma^2 \| \bbeta_* \|_2^2  \left( m(-\gamma) + \gamma m'(-\gamma ) \right) + \sigma^2 c \left( 1 - 2 \gamma m(-\gamma) - \gamma^2 m'(-\gamma ) \right), \\ 
  R_{{\rm out}, n \sim p}(\gamma) &= - \gamma^2 \| \bbeta_* \|_2^2  m'(-\gamma) + \sigma^2 c \left( m(-\gamma) + \gamma m'(-\gamma ) \right),
\end{align*}
where $m'(z)$ is the derivative of the Stieltjes transform $m(z)$ defined in the Mar{\u c}enko-Pastur equation in \Cref{eq:MP} with respect to $z$, so that
\begin{equation}
  m'(z) = \frac{ m(z) (c m(z) + 1 ) }{ -2 c z m(z) + 1 - c - z }.
\end{equation}

\paragraph{Ridgeless case.}
In the ridgeless setting as $\gamma \to 0$, one has $m(\gamma = 0) = \frac1{1-c} > 0$ only if $c < 1$ (that is, in the over-determined regime with $n > p$) and $\lim_{\gamma \to 0} m(-\gamma)$ is undefined otherwise, but satisfying $\lim_{\gamma \to 0} \gamma m(-\gamma) = \frac{c-1}c > 0$, in the under-determined regime with $n < p$. 
As such, one has
\begin{equation}\label{eq:LS_ridgeless_c<1}
  R_{\rm in}(\bbeta_0) \to \sigma^2 c, \quad R_{\rm out}(\bbeta_0) \to \sigma^2 \frac{c}{1-c},
\end{equation}
for $c<1$, and
\begin{equation}\label{eq:LS_ridgeless_c>1}
  R_{\rm out}(\bbeta_0) - \| \bbeta_* \|_2^2 \left(1- \frac1c \right) - \sigma^2 \frac1{c - 1} \to 0,
\end{equation}
for $c > 1$. 
This concludes the proof of \Cref{prop:risk_LS} in the proportional $n \sim p \to \infty$ limit.

\section{Proof of \Cref{rem:scaling_train_NN}}
\label{sec:proof_of_rem:scaling_train_NN}

It follows from \Cref{theo:nonlinear-Gram} that in the proportional regime for $n,p,d$ all large and $d < n$ (i.e., in the under-parameterized regime) and full rank $\K$, that $\delta$ \emph{diverges} as $\gamma\to 0$, but
\begin{equation}
  \gamma \delta = \frac1n \tr \K \left( \frac{d}n \frac{\K}{\gamma + \gamma \delta} + \I_n \right)^{-1} \xrightarrow{\gamma \to 0} \theta = \frac1n \tr \K \left( \frac{d}n \frac{\K}{\theta} + \I_n \right)^{-1},
\end{equation}
remains bounded as $\gamma \to 0$, so that both $\delta$ and $\tilde \Q$ scale like $1/\gamma$ as $\gamma \to 0$, and $\theta$ is the solution to
\begin{equation}\label{eq:def_theta}
  \frac{d}n = \int \frac{t}{t + \frac{n}d \theta} \mu_{\K}(dt),
\end{equation}
with $\mu_{\K}(dt)$ the ESD of $\K$ as in \Cref{def:ESD}.

We have in particular $\gamma \tilde \Q \xrightarrow{\gamma \to 0} \frac{n}d \theta \left( \K  + \frac{n}d \theta \I_n \right)^{-1}$, with $\frac{n-d}{n \theta} = \frac1d \tr \left( \K + \frac{n}d \theta \I_n \right)^{-1}$.
As a consequence, it follows from \Cref{prop:NN-MSEs} that for the training MSE in \Cref{def:single-layer-NN} that
\begin{equation}
  \tilde E_{\train} \xrightarrow{\gamma \to 0} \frac1n \y^\top \left( \frac{d}n \frac{\K}{\theta} +  \I_n \right)^{-1} \y,
\end{equation}
for any given training data $\X$ and target $\y$.
In particular, for $\y$ satisfying 
\begin{equation}
  \y =  \K^{1/2} \bbeta_*,  \quad \bbeta_* \sim \NN(\zo, \I_n),
\end{equation}
we further obtain
\begin{equation}
  \tilde E_{\train} \xrightarrow{\gamma \to 0} \frac1n \y^\top \left( \frac{d}n \frac{\K}{\theta} +  \I_n \right)^{-1} \y \simeq \frac1n \tr  \K \left( \frac{d}n \frac{\K}{\theta} +  \I_n \right)^{-1} = \theta.
\end{equation}

Let $\lambda_1\geq\cdots\geq\lambda_n>0$ be the eigenvalues of $\K$, and set $\rho = \lim d/n$.
From the fixed-point equation above, $\theta$ is equivalently determined by
\begin{equation}
  \rho
  =
  \frac1n
  \sum_{j=1}^n
  \frac{\lambda_j}
  {\lambda_j+\theta/\rho}.
\end{equation}

Let us consider the case of \textbf{polynomial eigendecay} for the matrix $\K$ (this is the case for, e.g., Matérn-type kernels~\cite{rasmussenGaussianProcessesMachine2005}), that is 
\begin{equation}
  \lambda_j
  =
  Cj^{-(1+\beta)}(1+o(1)),
  \qquad \beta>0.
\end{equation}
Let $q=1+\beta$ and set $s=\theta/\rho$. 
We look for a solution of the form $s=Cb n^{-q}$. 
Substituting this ansatz into the fixed-point equation gives
\begin{equation}
  \rho
  =
  \frac1n
  \sum_{j=1}^n
  \frac{1+o(1)}
  {1+b(j/n)^q+o(1)}.
\end{equation}
If $d/n\to\rho\in(0,1)$, the monotonicity imply that $b\to b_{\rho}$, where $b_{\rho}>0$ is the unique solution of
\begin{equation}
  \rho
  =
  \int_0^1
  \frac{dx}{1+b_{\rho}x^{1+\beta}}.
\end{equation}
Consequently,
\begin{equation}
  \theta
  =
  \rho s
  \sim
  C\rho\, b_{\rho} n^{-1-\beta}.
\end{equation}
This concludes the proof of \Cref{rem:scaling_train_NN}.

\section{Proof of \Cref{theo:linearization_kernel}}
\label{sec:proof_of_theo:linearization_kernel}

From its definition in \eqref{eq:def_K_NN}, the $(i,j)$ entry of $\K$ is given, for Gaussian $\w \sim \NN(\zo,\I_p)$, $\mathbf{x}_1, \ldots, \mathbf{x}_n \overset{\text{i.i.d.}}{\sim} \mathcal{U}(\mathbb{S}^{p-1})$ independently and uniformly drawn from the unit sphere in $\RR^p$, and $i \neq j$, by
\begin{equation*}
  [\K]_{ij} = \EE_\w[\phi(\x_i^\top \w) \phi(\w^\top \x_j)], 
\end{equation*}
for $(\x_i^\top \w,\x_j^\top \w) \sim \NN \left(\zo, \Big[\begin{smallmatrix} \| \x_i \|_2^2=1 & \x_i^\top \x_j \\ \x_i^\top \x_j & \| \x_j \|_2^2=1 \end{smallmatrix} \Big] \right)$.
Using the Gram-Schmidt orthogonalization procedure for standard Gaussian, we introduce 
\begin{equation}
  \x_i^\top \w = \xi_i, \quad \x_j^\top \w = (\x_i^\top \x_j) \cdot \xi_i + \sqrt{ 1 - (\x_i^\top \x_j)^2 } \cdot \xi_j,
\end{equation}
for independent standard Gaussian $\xi_i,\xi_j \sim \NN(0,1)$, so that 
\begin{align*}
  [\K]_{ij} &= \EE \left[ \phi(\xi_i) \phi \left( (\x_i^\top \x_j) \cdot \xi_i + \sqrt{ 1 - (\x_i^\top \x_j)^2 } \cdot \xi_
  j \right) \right] \\ 
  & = \EE \left[ \phi(\xi_i) \phi \left( \varepsilon_{ij} \xi_i + \left(1 - \frac{\varepsilon_{ij}^2}2 + O(\varepsilon_{ij}^4) \right) \xi_j \right) \right] \\ 
  & = \EE[\phi(\xi_i) \phi (\xi_j)] + \EE \left[ \phi(\xi_i) \phi'(\xi_j) \left( \varepsilon_{ij} \xi_i - \frac{\varepsilon_{ij}^2}2 \xi_j \right) \right] + \frac12 \EE \left[ \phi(\xi_i) \phi''(\xi_j) \cdot \varepsilon_{ij}^2 \xi_i^2 \right] + O(\varepsilon_{ij}^3) \\ 
  & = a_{\phi;0}^2 + a_{\phi;1}^2 \varepsilon_{ij} + a_{\phi;2}^2 \varepsilon_{ij}^2  + O(\varepsilon_{ij}^3) ,
\end{align*}
by Taylor expansion, for $\x_i^\top \x_j = \varepsilon_{ij} \ll 1$, where we recall $a_{\phi;0}, a_{\phi;1}, a_{\phi;2}, \nu_{\phi}$ the Hermite coefficients of $\phi$ as defined in \Cref{theo:normalized_Hermite}.

Similarly, for the $i^{th}$ diagonal entry of $\K$, we have
\begin{equation}
  [\K]_{ii} = \EE_\w[\phi(\x_i^\top \w) \phi(\w^\top \x_i)] = \EE[\phi^2(\xi_i)] = \nu_{\phi}.
\end{equation}

Since $\x_1, \ldots, \x_n $ are random vectors independently and uniformly drawn from the unit sphere $\mathbb{S}^{p-1}$, then $\x_i^\top \x_j = \varepsilon_{ij} $ satisfies that $(\varepsilon_{ij} + 1)/2 \sim {\rm Beta}(\frac{p-1}2, \frac{p-1}2)$ and $\varepsilon_{ij} = O(p^{-1/2})$ with high probability, so that
\begin{equation}
  \K = a_{\phi;0}^2 \one_n \one_n^\top + a_{\phi;1}^2 \X^\top \X + a_{\phi;2}^2 (\X^\top \X)^{\circ 2} + \left(\nu_\phi - \sum_{i=0}^2 a_{\phi;i}^2 \right) \I_n + O_{\| \cdot \|_2}(n p^{-3/2}).
\end{equation}
This, together with the fact that $(\X^\top \X)^{\circ 2} = \frac1p \one_n \one_n^\top + \I_n + O_{\| \cdot \|_2}(p^{-1/2})$ from \cite{couillet2016kernel,kammounCovarianceDiscriminativePower2023}, concludes the proof of \Cref{theo:linearization_kernel}.

\section{Proof of \Cref{theo:CK}}
\label{sec:proof_of_theo:CK}

To prove \Cref{theo:CK}, consider the following recursive relation on the CK matrices, as a function of the depth $\ell \in \{1, \ldots, L \}$~\cite{jacot2018neural}:
\begin{equation}\label{eq:recursion_CK}
  [\K_\ell]_{ij} = \EE_{(u,v)}[ \phi_\ell(u) \phi_\ell(v) ], \quad \K_0 = \X^\top \X, \quad (u,v) \sim \NN \left( \zo_2, \begin{bmatrix} [\K_{\ell-1}]_{ii} & [\K_{\ell-1}]_{ij} \\ [\K_{\ell-1}]_{ij} & [\K_{\ell-1}]_{jj} \end{bmatrix} \right).
\end{equation}
Then, following the idea of the proof of \Cref{theo:linearization_kernel} in \Cref{sec:proof_of_theo:linearization_kernel}, we can iteratively expand the nonlinear activation $\phi_\ell$ at layer $\ell$ using \Cref{theo:normalized_Hermite}.

For the case $\ell = 1$, it follows from \Cref{sec:proof_of_theo:linearization_kernel} that for $a_{\phi_1;0} = 0$ and $\nu_{\phi_1} = 1$, we have
\begin{equation}
  \K_1 = a_{\phi;1}^2 \X^\top \X + a_{\phi;2}^2 \cdot \frac1p \one_n \one_n^\top + \left(1 - a_{\phi;1}^2 \right) \I_n +  O_{\| \cdot \|_2}(p^{-1/2}).
\end{equation}
This satisfies the recursion in \Cref{theo:CK} with
\begin{equation}
  \alpha_{\ell=1,1} = a_{\phi_1;1}, \quad \alpha_{\ell=1,2} = a_{\phi_1;2} = \sqrt{ a_{\phi_1;1}^2 \cdot \alpha_{\ell=0,2}^2 + a_{\phi_1;2}^2 \cdot \alpha_{\ell=0,1}^2 },
\end{equation}
for $\alpha_{\ell=0,1} = 1$ and $\alpha_{\ell=0,2} = 0$, as a consequence of the fact that $\K_0 = \X^\top \X$.

We then prove \Cref{theo:CK} by induction on $\ell \in \{1, \ldots L\}$.
Assume that for layer $\ell -1$, we have
\begin{equation}\label{eq:K_ell-1_ij}
  [\K_{\ell-1}]_{ij} = \alpha_{\ell-1,1}^2 \x_i^\top \x_j + \alpha_{\ell-1,2}^2 (\x_i^\top \x_j)^2 + O(p^{-3/2}),
\end{equation}
for $i \neq j$ and 
\begin{equation}\label{eq:K_ell-1_ii}
  [\K_{\ell-1}]_{ii} = 1 + O(p^{-3/2}),
\end{equation}
which holds for $\ell = 2$.
Then, using the Gram-Schmidt orthogonalization procedure for standard Gaussian, we write, for $(u,v)$ defined in \Cref{eq:recursion_CK} that
\begin{align*}
  u &= \sqrt{ [\K_{\ell-1}]_{ii} } \cdot \xi_i = \xi_i + O(p^{-3/2}), \\ 
  v &= \frac{ [\K_{\ell-1}]_{ij} }{\sqrt{[\K_{\ell-1}]_{ii}}} \cdot \xi_i + \sqrt{ [\K_{\ell-1}]_{jj} - \frac{[\K_{\ell-1}]_{ij}^2}{[\K_{\ell-1}]_{ii}} } \cdot \xi_j = [\K_{\ell-1}]_{ij} \cdot \xi_i + \sqrt{ 1 - [\K_{\ell-1}]_{ij}^2} \cdot \xi_j + O(p^{-3/2}),
\end{align*}
for \emph{independent} $\xi_i, \xi_j \sim \NN(0,1)$.
Then, by \Cref{eq:K_ell-1_ij} and \Cref{eq:K_ell-1_ii} and Taylor expansion, we further get
\begin{align*}
  v &= \left( \alpha_{\ell-1,1}^2 \x_i^\top \x_j + \alpha_{\ell-1,2}^2 (\x_i^\top \x_j)^2 \right) \cdot \xi_i + \sqrt{ 1 -  \left( \alpha_{\ell-1,1}^2 \x_i^\top \x_j + \alpha_{\ell-1,2}^2 (\x_i^\top \x_j)^2 \right)^2 } \cdot \xi_j + O(p^{-3/2}) \\ 
  &= \left( \alpha_{\ell-1,1}^2 \x_i^\top \x_j + \alpha_{\ell-1,2}^2 (\x_i^\top \x_j)^2 \right) \cdot \xi_i + \left(1 - \frac12 \alpha_{\ell-1,1}^4 (\x_i^\top \x_j)^2 + O(p^{-2}) \right) \cdot \xi_j + O(p^{-3/2}) \\
  &=  \xi_j + \left( \alpha_{\ell-1,1}^2 \x_i^\top \x_j + \alpha_{\ell-1,2}^2 (\x_i^\top \x_j)^2 \right) \cdot \xi_i - \frac12 \alpha_{\ell-1,1}^4 (\x_i^\top \x_j)^2 \cdot \xi_j + O(p^{-3/2}),
\end{align*}
where we recall (as in \Cref{sec:proof_of_theo:linearization_kernel} for the proof of \Cref{theo:linearization_kernel}) that for $\mathbf{x}_1, \ldots, \mathbf{x}_n \overset{\text{i.i.d.}}{\sim} \mathcal{U}(\mathbb{S}^{p-1})$ independently and uniformly drawn from the unit sphere in $\RR^p$, we have $\x_i^\top \x_j = O(p^{-1/2})$ with high probability.

By this approximation, we then get from \Cref{eq:recursion_CK} that
\begin{align*}
  &[\K_\ell]_{ij} = \EE_{(u,v)}[ \phi_\ell(u) \phi_\ell(v) ] \\ 
  &= \EE_{\xi_i,\xi_j} \left[ \phi_\ell(\xi_i) \phi_\ell \left( \xi_j + \left( \alpha_{\ell-1,1}^2 \x_i^\top \x_j + \alpha_{\ell-1,2}^2 (\x_i^\top \x_j)^2 \right) \cdot \xi_i - \frac12 \alpha_{\ell-1,1}^4 (\x_i^\top \x_j)^2 \cdot \xi_j \right) \right] + O(p^{-3/2}) \\ 
  &= \EE \left[ \phi_\ell(\xi_i) \cdot \left( \phi_\ell(\xi_j) + \phi_\ell'(\xi_j) \left( \left( \alpha_{\ell-1,1}^2 \x_i^\top \x_j + \alpha_{\ell-1,2}^2 (\x_i^\top \x_j)^2 \right) \cdot \xi_i - \frac12 \alpha_{\ell-1,1}^4 (\x_i^\top \x_j)^2 \cdot \xi_j \right) \right) \right] \\ 
  &+ \frac12 \EE \left[ \phi_\ell(\xi_i) \cdot \phi_\ell''(\xi_j) \cdot \xi_i^2 (\alpha_{\ell-1,1}^2 \x_i^\top \x_j)^2 \right] + O(p^{-3/2}) \\ 
  &= a_{\phi_\ell;0}^2 + \EE[ \phi_\ell(\xi_i) \xi_i] \EE[\phi_\ell'(\xi_j) ] \left( \alpha_{\ell-1,1}^2 \x_i^\top \x_j + \alpha_{\ell-1,2}^2 (\x_i^\top \x_j)^2 \right) + \frac12 \EE [ \phi_\ell(\xi_i) \xi_i^2] \cdot \EE[\phi_\ell''(\xi_j)] \cdot (\alpha_{\ell-1,1}^2 \x_i^\top \x_j)^2 \\ 
  &- \frac12 a_{\phi_\ell;0}  \alpha_{\ell-1,1}^4 (\x_i^\top \x_j)^2  \EE[\phi_\ell'(\xi_j) \xi_j] + O(p^{-3/2}) \\ 
  &= a_{\phi_\ell;1}^2 \cdot \alpha_{\ell-1,1}^2 \cdot \x_i^\top \x_j + \left( a_{\phi_\ell;1}^2 \cdot \alpha_{\ell-1,2}^2 + a_{\phi_\ell;2}^2 \cdot \alpha_{\ell-1,1}^4 \right) (\x_i^\top \x_j)^2  + O(p^{-3/2})
\end{align*}
again by Taylor expansion of $\phi_\ell$ around $\xi_j$, where we exploited the independence between $\xi_i$ and $\xi_j$ in the fourth equality, and the assumption that $a_{\phi_\ell;0} = 0$ for all $\ell$, as well as the Stein's lemma in the fifth equality to yield
\begin{equation}
  \EE[\phi_\ell'(\xi) ] = \EE[ \phi_\ell(\xi) \xi] = a_{\phi_\ell;1}, \quad \EE [ \phi_\ell(\xi) \xi^2] = \EE[\phi_\ell''(\xi)] = \sqrt{2} a_{\phi_\ell;2}, 
\end{equation}
for Hermite coefficients $a_{\phi_\ell;1}, a_{\phi_\ell;2}$ as defined in \Cref{theo:normalized_Hermite}.

Note that this agrees with the expression of $[\K_{\ell-1}]_{ij}$ given in \Cref{eq:K_ell-1_ij} with
\begin{equation}
  \alpha_{\ell,1}^2 = a_{\phi_\ell;1}^2 \cdot \alpha_{\ell-1,1}^2, \quad \alpha_{\ell,2}^2 = a_{\phi_\ell;1}^2 \cdot \alpha_{\ell-1,2}^2 + a_{\phi_\ell;2}^2 \cdot \alpha_{\ell-1,1}^4.
\end{equation}
Similarly, for the $i^{th}$ diagonal entry of $\K_\ell$, we have
\begin{equation}
  [\K_\ell]_{ii} = \EE_u[\phi_\ell^2(u)] = \EE[\phi_\ell^2(\xi_i)] = \nu_{\phi_\ell} = 1.
\end{equation}
Putting these approximations in matrix form as in proof of \Cref{theo:linearization_kernel} in \Cref{sec:proof_of_theo:linearization_kernel}, and using the fact that $(\X^\top \X)^{\circ 2} = \frac1p \one_n \one_n^\top + \I_n + O_{\| \cdot \|_2}(p^{-1/2})$ from \cite{couillet2016kernel,kammounCovarianceDiscriminativePower2023}, we concludes the proof of \Cref{theo:CK}.
\fi

\end{document}